\documentclass[review]{elsarticle}

\usepackage{lineno,hyperref}
\usepackage{balance}
\usepackage{makecell}
\usepackage{threeparttable}  
\usepackage{adjustbox}
\usepackage{graphicx}
\usepackage{subcaption}

\usepackage{amssymb}
\usepackage{amsfonts}
\usepackage{multirow}
\usepackage{graphicx}
\usepackage{booktabs}
\usepackage{adjustbox}
\usepackage{array}
\usepackage{hyperref}

\usepackage{subcaption}

\usepackage{amsmath}

\usepackage{amsmath,amssymb,amsfonts,bm}

\usepackage{multirow}
\usepackage{tablefootnote}

\usepackage{color}
\usepackage{algorithm}
\usepackage{algorithmic}

\usepackage{multicol}
\usepackage{booktabs}
\usepackage{appendix}

\journal{Journal of \LaTeX\ Templates}

\begin{document}
\sloppy
\begin{frontmatter}

\title{Incorporating Attributes and Multi-Scale Structures for Heterogeneous Graph Contrastive Learning}

\author{Ruobing Jiang}
\ead{jrb@ouc.edu.cn}

\author{Yacong Li}
\ead{liyacong1@stu.ouc.edu.cn}

\author{Haobing Liu\corref{corresponding}}
\ead{haobingliu@ouc.edu.cn}

\author{Yanwei Yu}
\ead{yuyanwei@ouc.edu.cn}

\fntext[equal]{Yacong Li contributed equally to this work as the first author.}
\address{Department of Computer Science and Technology, Ocean University of China}
\tnotetext[accept]{This paper has been accepted for publication in \textit{Information fusion}.}

\cortext[corresponding]{Corresponding author.}

\begin{abstract}
Heterogeneous graphs (HGs) are composed of multiple types of nodes and edges, making it more effective in capturing the complex relational structures inherent in the real world. However, in real-world scenarios, labeled data is often difficult to obtain, which limits the applicability of semi-supervised approaches. Self-supervised learning aims to enable models to automatically learn useful features from data, effectively addressing the challenge of limited labeling data. In this paper, we propose a novel contrastive learning framework for heterogeneous graphs (ASHGCL), which incorporates three distinct views, each focusing on node attributes, high-order and low-order structural information, respectively, to effectively capture attribute information, high-order structures, and low-order structures for node representation learning. Furthermore, we introduce an attribute-enhanced positive sample selection strategy that combines both structural information and attribute information, effectively addressing the issue of sampling bias. Extensive experiments on four real-world datasets show that ASHGCL outperforms state-of-the-art unsupervised baselines and even surpasses some supervised benchmarks.
\end{abstract}


\end{frontmatter}


\section{Introduction}

Heterogeneous graphs (HGs)~\cite{HG} are composed of various types of nodes or/and edges, making them more effective than homogeneous graphs in capturing the complex relational structures inherent in the real world~\cite{thanho}. Heterogeneous graphs have been widely applied in various domains, including academic networks~\cite{academic}, recommendation systems~\cite{recommendation} and biomedical research~\cite{FCHGNN,DRMAHGC}. For example, in academic networks, entities such as authors, papers, and conferences can form a heterogeneous graph. Heterogeneous graphs enable the discovery of deeper associations by capturing the semantics of multi-typed nodes and edges, which provide more informative node representations and contextual features for tasks such as node classification~\cite{HAN} and link prediction~\cite{GATNE}, underscoring its significance in practical applications.

Traditional heterogeneous graph representation learning methods~\cite{HAN,MHGCN,MAGNN,HPN,MECCH} primarily rely on semi-supervised learning. These methods use a small amount of labeled data to guide the model in learning node representations. However, in real-world scenarios, labeled data is often difficult to obtain, which limits the applicability of semi-supervised approaches. Self-supervised learning~\cite{SLL} aims to enable models to learn representations from raw data without relying on labels, effectively addressing the challenge of label scarcity. As a key unsupervised approach, clustering has been explored in multi-view settings~\cite{cluster1, cluster2} to capture shared and complementary information for representation learning. Among self-supervised techniques, contrastive learning~\cite{CL1,CL2} has emerged as a widely adopted paradigm. The core idea of contrastive learning is to bring similar instances closer and push dissimilar ones apart, thereby improving the quality of representations. Recent studies have begun to explore the application of contrastive learning techniques in heterogeneous graph representation learning~\cite{HeCo,HGCML,HGCMA,HeMuc,STENCIL,ESHG}.

\begin{table}
\centering
\caption{Comparison of view settings between our proposed model and other heterogeneous graph contrastive learning models.}
\label{viewpk}
\begin{tabular}{lccc}
\hline
Model & high-order\textsuperscript{1} & low-order\textsuperscript{2} &  attribute\textsuperscript{3} \\
\hline
HeCo~\cite{HeCo} & $\checkmark$ & $\checkmark$ & $\times$ \\
HGCML~\cite{HGCML}  & $\checkmark$ & $\times$ & $\times$ \\
HGCMA~\cite{HGCMA}  & $\checkmark$ & $\times$ & $\times$ \\
HeMuc~\cite{HeMuc}  & $\checkmark$ & $\times$ & $\checkmark$ \\
STENCIL~\cite{STENCIL}  & $\checkmark$ & $\times$ & $\times$ \\
MHGCL~\cite{MHGCL}& $\checkmark$ & $\checkmark$ & $\times$ \\
MEOW~\cite{MEOW} & $\checkmark$ & $\times$ & $\times$ \\
our proposed model & $\checkmark$ & $\checkmark$ & $\checkmark$ \\
\hline
\end{tabular}
\begin{tablenotes}
\small
\item1: “high-order” means that it captures high-order structural information.
\item2: “low-order” means that it captures low-order structural information.
\item3: “attribute” means that it leverages semantic information inherent in node attributes.
\end{tablenotes}
\end{table}

\begin{figure}[!t]
	\centering
	\includegraphics[width=2.5in]{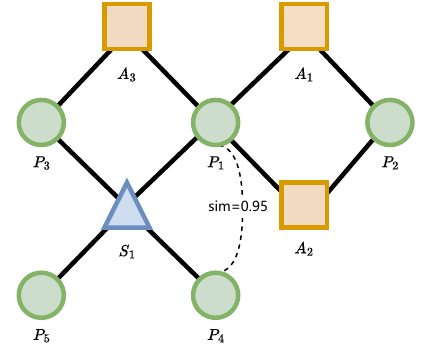}
	\caption{An example of sampling bias.}
	\label{bias}
\end{figure}

In the realm of heterogeneous graph contrastive learning methods, most existing methods such as HGCML~\cite{HGCML} and HGCMA~\cite{HGCMA} generate multiple views primarily through the utilization of meta-paths, focusing on capturing high-order structural information. Although these methods incorporate node attributes during the Graph Convolutional Network(GCN) aggregation process, they merely use attributes as initial node features without fully exploiting the rich information contained in attributes. Recent work like HeMuc~\cite{HeMuc} explores a different direction by introducing similarity-based views through attribute information. However, it overlooks the one-hop neighbors of target nodes, thereby compromising the completeness of the learned representations. Table~\ref{viewpk} shows the comparison between our proposed model and SOTA methods in terms of three important types of information utilized in their view settings. Moreover, existing methods like HeCo~\cite{HeCo} sample positive pairs based on the direct count of meta-paths between two nodes. This sampling strategy focuses exclusively on structural information, neglecting that nodes with similar attributes should also be treated as positive pairs. As shown in Figure~\ref{bias}, we consider the meta-paths PAP and PSP to capture structural relationships between nodes. If we only consider the structural similarity to the positive pairs of the sample for $P_1$, nodes $P_2$ and $P_3$ both share two meta-paths connected to $P_1$. As a result, they would be identified as positive samples for $P_1$. However, $P_4$, which is structurally far from $P_1$ with only one meta-path connecting them, shares highly similar attributes with $P_1$. Therefore, $P_4$ should also be considered as a positive sample during the sampling process.

It is challenging to design an efficient heterogeneous graph contrastive learning method, as it requires addressing the following aspects:

\textbf{(1) How to fully leverage node attributes, low-order, and high-order structural information when constructing views?} For self-supervised learning methods, the design of different views plays a critical role in learning effective representations~\cite{view}. Node attributes inherently carry rich semantic information~\cite{feature1,feature2}, and nodes with similar attributes are likely to belong to the same category. Most methods~\cite{HeCo,HGCML,HGCMA,MHGCL} fail to simultaneously capture both node attributes and structural information, while methods like HeMuc~\cite{HeMuc} struggle to effectively integrate high-order and low-order structures. To address the challenge, we construct multiple views that integrate both node attributes and multi-scale structural information. Specifically, we construct three distinct views based on low-order structural information, high-order structural information, and node attribute information, respectively, to fully exploit the multiple types of information inherent in heterogeneous graphs. 

\textbf{(2) How to effectively leverage graph structure and node attribute semantics during the sampling process?} The core idea of contrastive learning is to pull similar samples closer, making effective sampling crucial. Existing methods typically rely on structural information for node sampling, either by assessing node proximity through the number of meta-paths between node pairs~\cite{HeCo} or using Personalized PageRank (PPR) scores as a proximity measure to determine the sampling probability of node pairs~\cite{STENCIL}. However, these approaches frequently overlook that nodes with similar attributes should also be treated as positive pairs. They often suffer from sampling bias~\cite{samplebias}, where nodes with rich structural connections are oversampled while attribute-similar but structurally-distant node pairs are undersampled. To address the challenge, we propose an attribute-enhanced positive sampling strategy that samples positive pairs based on both structural information and node attribute information, effectively mitigating sampling bias and improving the overall quality of learned representations.

In summary, we propose a novel Heterogeneous Graph Contrastive Learning model which incorporates Attributes and multi-scale Structure (ASHGCL in short). This framework leverages three distinct views: a feature similarity view that mines inherent relationships embedded in node attributes by computing attribute similarities, a high-order relation view based on meta-paths, and a low-order relation view utilizing first-order neighbors. This multi-view design captures comprehensive attribute, low-order, and high-order structural information. Furthermore, we introduce an attribute-enhanced positive sample selection strategy that combines both structural and attribute information, which significantly enhances the quality of positive samples and optimizes the overall contrastive learning process. Additionally, building on the commonly used local contrast that performs node-level comparison between positive pairs, we introduce a global contrastive mechanism to effectively capture global knowledge from the views, further enriching the node embeddings. In summary, the main contributions of this work are as follows:

$\bullet$ We propose a novel contrastive learning framework for heterogeneous graphs, ASHGCL, which incorporates three distinct views to effectively capture attribute information, high-order structures, and low-order structures for node representation learning.

$\bullet$ We introduce an attribute-enhanced positive sampling strategy that leverages both structural and attribute information to select high-quality positive samples for each anchor node, effectively addressing the challenge of sampling bias.

$\bullet$ We apply local and global contrast to capture knowledge at both the node and graph levels. It enhances the richness of node embeddings at both fine-grained and coarse-grained levels, providing them with stronger representational capabilities.

$\bullet$ Extensive empirical evaluations are carried out on four real-world datasets, demonstrating that ASHGCL not only exceeds state-of-the-art unsupervised baselines, but also exceeds the performance of certain supervised baselines.

The remainder of this paper is organized as follows. \autoref{rw} introduces the related work. \autoref{pre} outlines the preliminaries. \autoref{mo} describes the proposed model. \autoref{ex} presents the experiments and results. Finally, \autoref{co} concludes the paper.

\section{Related Work}\label{rw}

In this section, we briefly review two categories of related work: heterogeneous graph neural networks and graph contrastive learning.
\subsection{Heterogeneous Graph Neural Network}

Graph Neural Networks (GNNs) have been widely applied to tasks involving homogeneous graphs~\cite{hom1,hom2,hom3,high-order}, where all nodes and edges are of the same type. In recent years, some researchers have started to extend GNNs to heterogeneous graphs. For example, HAN~\cite{HAN} employs a dual-level attention mechanism to aggregate information from the endpoints of meta-paths. MAGNN~\cite{MAGNN} extends this approach by incorporating intermediate nodes along the meta-paths into the aggregation process. MECCH~\cite{MECCH} introduces metapath context to prevent information loss or redundancy during aggregation. SeHGNN~\cite{SeHGNN} argues that using attention mechanisms for intra-meta-path aggregation is not necessary and proposes using simple mean aggregation for inter-meta-path fusion. ie-HGCN~\cite{iehgnn} also avoids using attention within meta-paths and proposes an efficient heterogeneous graph convolutional network, significantly enhancing model performance. GTN~\cite{GTN} implements automatic meta-path generation, eliminating the need for manual meta-path design. 

However, these supervised learning methods~\cite{HAN,MAGNN,SeHGNN} require labeled data to guide the learning process. In practical applications, acquiring a substantial volume of labeled data is often challenging and even unfeasible~\cite{SHGP}.

\subsection{Graph Contrastive Learning}

Graph contrastive learning~\cite{MVGRL} applies the principles of contrastive learning to graph neural networks, mainly by constructing contrastive objectives based on different graph structures or augmentations, thus enhancing the effectiveness of graph representation learning. Graph contrastive learning has been widely applied to homogeneous graphs. Specifically, DGI~\cite{DGI} maximizes mutual information between local patch representations and global graph summaries, using the InfoMax~\cite{Infomax} principle to construct an objective. MVGRL~\cite{MVGRL} performs contrastive learning on views of first-order neighbors and graph diffusion. GMI~\cite{GMI} maximizes mutual information between input and node representations at both the attribute and structural levels. GRACE~\cite{GRACE} performs contrastive learning at the node level instead of at the graph level.

Recently, researchers have begun exploring the extension of contrastive learning to heterogeneous graph ~\cite{HeCo,HeMuc,HGCMA,HGCML,STENCIL,MDHGCL}. DMGI~\cite{DMGI} adapts DGI for heterogeneous networks using meta-path based views and integrates embeddings via consensus regularization. HeCo~\cite{HeCo} sets up two distinct views, meta-path and network schema, enabling the simultaneous capture of both high-order and low-order structural information. STENCIL~\cite{STENCIL} treats each meta-path as a separate view and proposes a structure-enhanced negative sampling strategy. HGCML\cite{HGCML} employs Personalized PageRank (PPR) and L2 distance as metrics to simultaneously capture both structural and attribute positive samples. HeMuc\cite{HeMuc} establishes a reachability matrix to capture high-order structural information and constructs contrastive views based on node feature similarity. BPHGNN~\cite{BPHGNN} introduces deep behavior patterns and broad behavior patterns and improves the consistency of node representations in different views through contrastive learning.

However, these approaches either fail to simultaneously leverage both node attributes and graph structural information, or they do not adequately account for both high-order structures and low-order structures, limiting the expressiveness of the models.

\section{Preliminary}\label{pre}

In this section, we introduce some basic concepts related to heterogeneous graphs, as well as the specific tasks we aim to address.

\begin{figure}[!tb]
	\centering
	\includegraphics[width=3.4in]{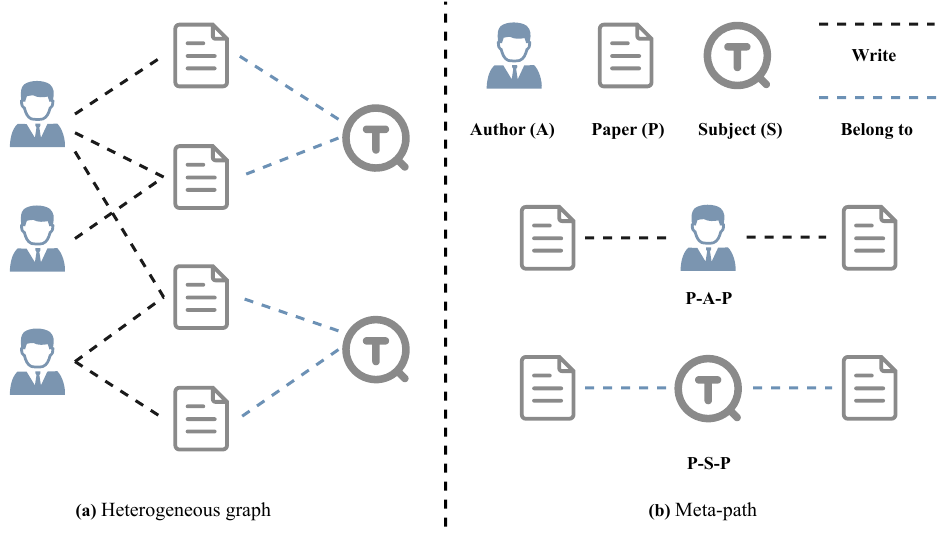}
	\caption{A heterogeneous graph example and its meta-path illustrations.}
	\label{HG}
\end{figure}

\textbf{Definition 1. Heterogeneous Graphs.} A heterogeneous graph is defined as a graph \( G = (V, E, \phi, \psi) \), where \( V \) and \( E \) represent the sets of nodes and edges, respectively. The mapping functions \( \phi(v) : V \rightarrow O \) and \( \psi(e) : E \rightarrow R \) assign each node \( v \in V \) to a node type from \( O \) and each edge \( e = (v, v') \in E \) to an edge type from \( R \). The condition \( |O| + |R| > 2 \) ensures the graph contains more than one type of node or edge.

For example, Figure~\ref{HG}(a) is an instance of a heterogeneous graph that contains three types of nodes: paper, author, and subject. The graph contains two edge types: one representing author writes paper, and the other denoting paper belongs to subject.

\textbf{Definition 2. Meta-path.} A meta-path \( P \) is defined as
\[
P = O_1 \xrightarrow{R_1} O_2 \xrightarrow{R_2} \ldots \xrightarrow{R_l} O_{l+1}
\]
(simplified as \( O_1O_2 \ldots O_{l+1} \)), where node types \( O_1, O_2, \ldots, O_{l+1} \in O \) and edge types \( R_1, R_2, \ldots, R_l \in R \).

For example, in Figure~\ref{HG}(b), the meta-path PAP indicates that two papers are written by the same author, while the meta-path PSP represents that two papers belong to the same subject. Meta-paths can capture high-order neighbors of nodes, thereby providing richer structural and node attribute information.

\textbf{Problem Definition.} Heterogeneous graph representation learning aims to learn low-dimensional vector embeddings for nodes from a given heterogeneous graph $G = (V, E, \phi, \psi)$. Specifically, this task maps the nodes to a lower-dimensional space through a mapping function $f: V \rightarrow \mathbb{R}^d$, where $d \ll |V|$. The goal is to make sure that the learned node embeddings can preserve as much structural and semantic information of the heterogeneous graph as possible. These embeddings can then be used as input for downstream tasks such as node classification, node clustering, and link prediction.

\section{The Proposed Model: ASHGCL}\label{mo}

\begin{figure*}[!tb]
	\centering
	\includegraphics[width=4.7in]{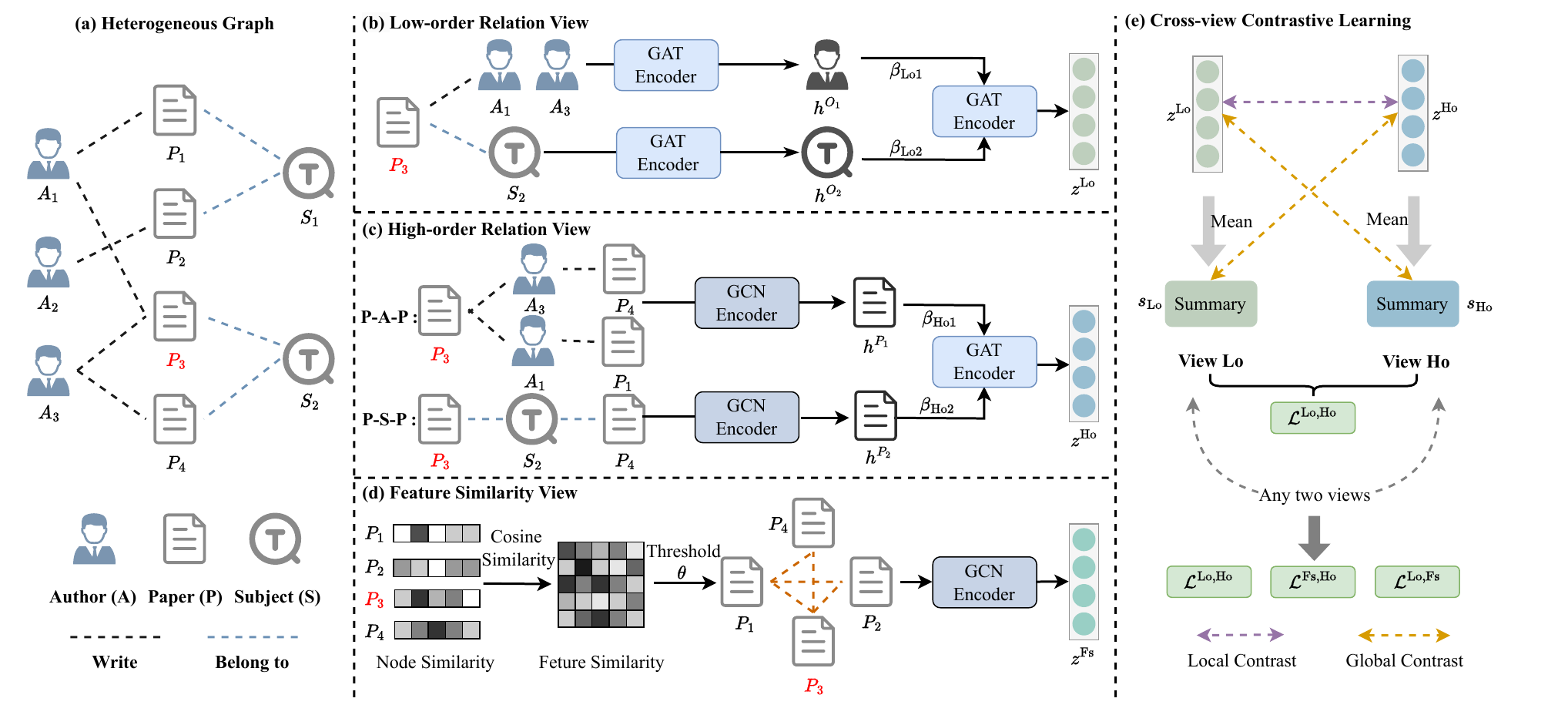}
	\caption{The overall framework of our proposed ASHGCL. (a) The original heterogeneous graph. (b) The low-order relation view, which aggregates first-order neighbors of target nodes using two-step GAT. (c) The high-order relation view, which aggregates meta-path based neighbors using GCN and combines different meta-paths using GAT. (d) The feature similarity view, which computes similarities between node attributes, generates a similarity graph, and aggregates information using GCN. (e) The cross-view contrastive learning, which generates graph-level representations from the node-level representations. It performs pairwise comparisons between three views, where each comparison involves both local contrast among node-level representations and global contrast between node-level and graph-level representations.}
	\label{ASHGCL}
\end{figure*}

In this section, we introduce ASHGCL, our proposed contrastive learning framework for heterogeneous graph representation learning. The overview architecture is illustrated in Figure~\ref{ASHGCL}. Our approach leverages three distinct views: a low-order relation view that captures immediate neighborhood structures, a high-order relation view that captures meta-path based relationships, and a feature similarity view that utilizes node attribute information. We introduce an attribute-enhanced positive sampling strategy that effectively integrates topological structure and node attributes, ensuring high-quality contrastive pairs for robust representation learning. Additionally, we apply local and global contrast to capture knowledge at both the node and graph levels.

\subsection{Feature Mask based Data Augmentation}

Before generating node embeddings using view-based encoders, we apply feature masking to node attributes as a form of data augmentation. This technique forces the model to infer missing attribute information, thereby increasing the learning challenge and encouraging the extraction of more discriminative node representations. As a result, the model learns to rely on more informative structural and contextual cues, enhancing its robustness against noise and missing data. Similar strategies have been shown to be effective in self-supervised learning frameworks for graphs~\cite{GraphMAE}. Notably, we do not perform feature masking on the feature similarity view, as its primary function is to capture semantic information inherent in node attributes. Applying random feature masking in this view would introduce noise and lead to significant loss of semantic content.

\subsection{Low-order Relation View}

The low-order relationship view focuses on the first-order neighbor information of nodes. For each node in a heterogeneous graph, we capture its low-order structural features by aggregating the features of its directly connected heterogeneous neighbor nodes. Specifically, for a given node $i$, we consider its connections with nodes of various types. Let $O = \{O_1, O_2, ..., O_M\}$ be the set of node types in the heterogeneous graph. For a node $i$, the neighbors of type $O_k$ can be denoted as $N^{O_k}_i$. We design a hierarchical attention mechanism that includes two steps: node-level attention and type-level attention. Specifically, we use node-level attention to compute the attention weights between node $i$ and its $O_k$-type neighbors:

\begin{equation}
\alpha^{O_k}_{ij} = \frac{\exp(\operatorname{LeakyReLU}(\mathbf{a}^ {\top}_{O_k} \cdot [h_i || h_j]))}{\sum_{l \in N^{O_k}_i} \exp(\operatorname{LeakyReLU}(\mathbf{a}^{\top}_{O_k} \cdot [h_i || h_l]))},
\end{equation}
where $\mathbf{a}^{\top}_{O_k}$is a learnable attention vector for type $O_k$, $h_i$ and $h_j$ are the feature vectors of nodes $i$ and $j$ respectively, and $||$ denotes the concatenation operation.Then we aggregate node $i$ with its $O_k$-type neighbors:

\begin{equation}
h^{O_k}_i = \sigma \left(\sum_{j \in N^{O_k}_i} \alpha^{O_k}_{ij}  h_j\right),
\end{equation}
where $\sigma$ is ELU activation function~\cite{ELU}. To enhance efficiency and reduce computational complexity, we randomly sample a fixed number of neighbors from each node type, rather than aggregating information from all neighbors.

Through node-level attention, we obtain the type-specific embeddings $h^{O_1}_i, h^{O_2}_i, ..., h^{O_M}_i$ for node $i$. Then we proceed to implement type-level attention to aggregate information across these different node type embeddings. First, we compute a type-specific weight for each node type:

\begin{equation}
w_{O_k} = \frac{1}{|V|} \sum_{i \in V} \mathbf{a}^{\top}_\text{Lo} \cdot \tanh(W_\text{Lo}h^{O_k}_i + b_\text{Lo}),
\end{equation}
where $\mathbf{a}^{\top}_\text{Lo}$, $W_\text{Lo}$, and $b_\text{Lo}$ are learnable parameters, and $|V|$ is the number of nodes in the graph.

To obtain the low-order relation view representation, we employ an attention mechanism to aggregate type-specific embeddings. The final low-order relation view representation for node $i$ is computed as:

\begin{equation}
z^\text{Lo}_i = \sum_{k=1}^M \frac{\exp(w_{O_k})}{\sum_{m=1}^M \exp(w_{O_m})} \cdot h^{O_k}_i,
\end{equation}
where $w_{O_k}$ is the importance weight for node type $O_k$, $M$ is the number of node types, and $h^{O_k}_i$ is the type-specific embedding of node $i$ for type $O_k$.

\subsection{High-order Relation View}

The high-order relation view captures complex, indirect relationships between nodes through meta-paths in heterogeneous graphs. We propose a dual-layer aggregation architecture (intra-meta-path and inter-meta-path) to effectively model these high-order relations. Let $P=\{P_1,P_2,...,P_L\}$ denote a set of predefined meta-paths. For each meta-path $P_l$, we use an enhanced graph convolution method combining neighborhood aggregation and degree normalization. Please note that in different structural views, we adopt different encoders based on computational efficiency and modeling capability. Specifically, in the low-order structure view, where the number of first-order neighbors is relatively small, we employ GAT to capture their importance through attention mechanisms. In contrast, in the high-order structure view, the number of meta-path-based neighbors is much larger, making GCN a more suitable choice for efficient aggregation. Prior studies~\cite{SeHGNN,iehgnn} have also shown that employing attention mechanisms within meta-paths yields negligible performance gains while significantly increasing computational costs. The representation of node $i$ under meta-path $P_l$ is:

\begin{equation}
h_i^{P_l} = \sigma\left(\frac{1}{d_i} \sum_{j \in N_i^{P_l}} h_j\right),
\end{equation}
where $N_i^{P_l}$ represents the meta-path based neighbors of node $i$ under $P_l$, $d_i$ is the degree of node $i$, and $\sigma$ is PReLU.

After obtaining embeddings for each meta-path, we employ an attention mechanism to aggregate information from different meta-paths. First, we calculate the importance weight for each meta-path:

\begin{equation}
w_{P_l} = \frac{1}{|V|} \sum_{i \in V} \mathbf{a}^{\top}_\text{Ho} \cdot \tanh(W_\text{Ho}h_i^{P_l} + b_\text{Ho}),
\end{equation}

Then, we use these weights to aggregate and obtain the high-order relation view representation for each node:

\begin{equation}
z_i^\text{Ho} = \sum_{l=1}^L \frac{\exp(w_{P_l})}{\sum_{m=1}^L \exp(w_{P_m})} \cdot h_i^{P_l},
\end{equation}
where $\mathbf{a}_{Ho}$, $W_{Ho}$, and $b_{Ho}$ are learnable parameters, and $L$ is the number of meta-paths. This high-order relation view enables our model to capture complex structural patterns and long-range dependencies in heterogeneous graphs, complementing the information obtained from the low-order relation view.

\subsection{Feature Similarity View}

The feature similarity view aims to capture the similarity between node attributes, thereby providing a complementary representation based on node features. First, we construct a similarity graph by computing the cosine similarity between pairs of nodes. For nodes $i$ and $j$, their similarity is defined as:

\begin{equation}
s_{ij} = \frac{h_i \cdot h_j}{\|h_i\| \|h_j\|},
\end{equation}
where $h_i$ and $h_j$ are the feature vectors of nodes $i$ and $j$ respectively, $\cdot$ denotes the dot product, and $\|\cdot\|$ represents the L2 norm of a vector. We use a threshold $\theta$ to determine the edges in the similarity graph: when $s_{ij} \geq \theta$, we set $a_{ij} = 1$, indicating an edge between nodes $i$ and $j$; otherwise, $a_{ij} = 0$, indicating no edge between nodes $i$ and $j$.

After constructing the feature similarity view, we employ a GCN to aggregate node information. The representation of node $i$ in the feature similarity view is computed as follows:

\begin{equation}
z_i^\text{Fs} = \sum_{j \in \mathcal{N}_i} W_\text{Fs} h_j,
\end{equation}
where $\mathcal{N}_i$ is the set of neighboring nodes of node $i$ in the feature similarity view, and $W_\text{Fs}$ is a learnable weight matrix.

\subsection{Attribute-enhanced Positive Sampling}

\textbf{Structural-aware Positive Sampling.} Existing methods often utilize Personalized PageRank (PPR) to quantify structural similarity between nodes. However, PPR, primarily based on random walk principles, tends to capture local neighborhood structures, potentially overlooking broader global structural features. In contrast, meta-paths consider various connection patterns rather than relying solely on random walk probabilities. Meta-paths describe connection patterns between different types of nodes in a graph, enabling the capture of high-order topological structures. We propose that if a pair of nodes is connected through multiple meta-paths, these nodes exhibit higher structural similarity.

We define the number of meta-paths between a pair of nodes $(i,j)$ as $\mathbb{C}(i,j)$:

\begin{equation}
\mathbb{C}(i,j) = \sum_{m=1}^{L} \mathbf{1}(j \in \mathcal{N}_i^{P_m}),
\end{equation}
where $\mathbf{1}(\cdot)$ is the indicator function, and $\mathcal{N}_i^{P_m}$ represents the set of neighboring nodes of node $i$ on the meta-path $P_m$. Then we select the top-\( k \) nodes with the highest meta-path counts as structural positive samples for each node, denoted as $\mathbb{P}^t$.

\textbf{Attribute-aware Positive Sampling.}
Many existing sampling strategies primarily rely on graph structure, with some even designating nodes from different views as positive samples, neglecting the rich semantic information embedded in node attributes. To effectively capture the attribute information of nodes, we utilize cosine similarity to measure the similarity between nodes. Similarly, we select the top-\( k \) nodes with the highest similarity to the anchor node as its attribute-aware positive samples, denoted as \( \mathbb{P}^s \).

\textbf{Sample Integration.}
Then we combine the two types of positive samples. The positive samples for node \(i\) are defined as:

\begin{equation}
\mathbb{P}_i = \mathbb{P}_i^t \cup \mathbb{P}_i^s,
\end{equation}
all other nodes not included in $\mathbb{P}_i$ are considered as negative samples, denoted as $\mathbb{N}_i$. The positive samples obtained through this method can integrate topological structure and node attributes, thereby enhancing the model's expressive capability.

\subsection{Cross-view Contrastive Learning}
After obtaining node representations from three views, we use an MLP with hidden layers to project these representations into a shared space:

\begin{equation}
\begin{aligned}
z_i^{\text{Fs}'} &= W^{(2)}(\sigma(W^{(1)}z_i^\text{Fs} + b^{(1)})) + b^{(2)}, \\
z_i^{\text{Lo}'} &= W^{(2)}(\sigma(W^{(1)}z_i^\text{Lo} + b^{(1)})) + b^{(2)}, \\
z_i^{\text{Ho}'} &= W^{(2)}(\sigma(W^{(1)}z_i^\text{Ho} + b^{(1)})) + b^{(2)},
\end{aligned}
\end{equation}
where $\sigma$ is a non-linear activation function, and the parameters $\{W^{(1)}, W^{(2)}, b^{(1)}, b^{(2)}\}$ are shared across all three views.

\textbf{Local Contrast.}
For each node $i$ in the graph, we define a set of positive samples $\mathbb{P}_i$ and negative samples $\mathbb{N}_i$. The local contrastive loss is defined as three terms, each corresponding to a pairwise comparison between two of the three views:

\begin{equation}
\begin{aligned}
\mathcal{L}^\text{Fs,Lo}_{l}(i) &= -\log \frac{\sum_{j \in \mathbb{P}_i} \exp(\text{sim}(z_i^{\text{Fs}'}, z_j^{\text{Lo}'}) / \tau)}{\sum_{k \in \{\mathbb{P}_i \cup \mathbb{N}_i\}} \exp(\text{sim}(z_i^{\text{Fs}'}, z_k^{\text{Lo}'}) / \tau)}, \\[10pt]
\mathcal{L}^\text{Lo,Ho}_{l}(i) &= -\log \frac{\sum_{j \in \mathbb{P}_i} \exp(\text{sim}(z_i^{\text{Lo}'}, z_j^{\text{Ho}'}) / \tau)}{\sum_{k \in \{\mathbb{P}_i \cup \mathbb{N}_i\}} \exp(\text{sim}(z_i^{\text{Lo}'}, z_k^{\text{Ho}'}) / \tau)}, \\[10pt]
\mathcal{L}^\text{Ho,Fs}_{l}(i) &= -\log \frac{\sum_{j \in \mathbb{P}_i} \exp(\text{sim}(z_i^{\text{Ho}'}, z_j^{\text{Fs}'}) / \tau)}{\sum_{k \in \{\mathbb{P}_i \cup \mathbb{N}_i\}} \exp(\text{sim}(z_i^{\text{Ho}'}, z_k^{\text{Fs}'}) / \tau)},
\end{aligned}
\end{equation}
where $\text{sim}(v, u)$ represents the cosine similarity between vectors $v$ and $u$, and $\tau$ is a temperature hyperparameter. Unlike traditional contrastive learning methods~\cite{1cons,1cons1} that typically use only one positive pair, our approach incorporates multiple positive samples in each loss term. 
The local objective is given as follows:

\begin{equation}
\mathcal{J}_{local} = \frac{1}{|V|} \sum_{i \in V} [\lambda_1 \cdot \mathcal{L}^\text{Fs,Lo}_{l}(i) + \lambda_2 \cdot \mathcal{L}^\text{Lo,Ho}_{l}(i) + \lambda_3 \cdot \mathcal{L}^\text{Ho,Fs}_{l}(i)],
\end{equation}
where $\lambda_1$, $\lambda_2$, and $\lambda_3$ are non-negative coefficients that balance the contributions of different contrastive loss terms.

\textbf{Global Contrast.}
In addition to employing local contrastive learning between node representations, we introduce a global contrastive mechanism that compares node representations with graph-level representations. This approach aids the model in capturing global structural information. The global contrastive objective is defined as follows:

\begin{equation}
\begin{aligned}
\mathcal{L}^{a,b}_{g}(i) &= -\log(D(z^{a'}_i, s_{a})) - \log(1 - D(z^{b'}_i, s_{a})),
\end{aligned}
\end{equation}
where $\mathcal{L}^{a,b}_{g}(i)$ denotes the global contrastive loss for node $i$ between views $a$ and $b$,  $s_a$ is the graph-level representation for view $a$ calculated via mean pooling, and $D(z, s) = \sigma(\text{BiLinear}(\rho(z), \rho(s)))$ is a discriminator, where $\rho(\cdot)$ is a projection function, BiLinear(·) is a bilinear layer, and $\sigma(\cdot)$ is the sigmoid function. Please note that $\mathcal{L}^{a,b}_{g}(i) \neq \mathcal{L}^{b,a}_{g}(i)$. The global objective is given as follows:

\begin{equation}
\begin{aligned}
\mathcal{J}_{global} = \frac{1}{|V|} \sum_{i \in V} [&\lambda_4 \cdot (\mathcal{L}^\text{Fs,Lo}_{g}(i) + \mathcal{L}^\text{Lo,Fs}_{g}(i)) + \\
&\lambda_5 \cdot (\mathcal{L}^\text{Lo,Ho}_{g}(i) + \mathcal{L}^\text{Ho,Lo}_{g}(i)) + \\
&\lambda_6 \cdot (\mathcal{L}^\text{Ho,Fs}_{g}(i) + \mathcal{L}^\text{Fs,Ho}_{g}(i))],
\end{aligned}
\end{equation}
where $\lambda_4$, $\lambda_5$, and $\lambda_6$ are non-negative coefficients that balance the contributions of different global contrastive loss terms.

\textbf{Overall Objective.} The overall objective is to combine the local and global contrastive losses. It is defined as:

\begin{equation}
\mathcal{J} = \mu \mathcal{J}_{local} + (1 - \mu) \mathcal{J}_{global}
\end{equation}
where $\mu$ is a hyperparameter that balances the contributions of the local and global losses. The local contrast primarily focuses on the fine-grained relationships between nodes, while the global contrastive mechanism captures global knowledge across three views. This combination enhances the richness of node embeddings.

\section{Experiments}\label{ex}

\subsection{ Experimental Setup}
\textbf{Datasets.}
We employ four real-world heterogeneous information network (HIN) datasets in our experiments. The basic information of these datasets is summarized in Table~\ref{tab:datasets}.
\begin{itemize}
    \item \textbf{ACM}~\cite{ACM}: It contains 4,019 papers, 7,167 authors, and 60 subjects. The target nodes are papers, which are classified into three categories.
    
    \item \textbf{DBLP}~\cite{MAGNN}: It contains 4,057 authors, 14,328 papers, 20 conferences, and 7,723 terms. The target nodes are authors, which are divided into four classes.
    
    \item \textbf{AMiner}~\cite{Aminer}: It contains 6,564 papers, 13,329 authors, and 35,890 references. The target nodes are papers, which are categorized into four classes.

    \item \textbf{Freebase}~\cite{Freebase}: It contains 3,492 movies, 33,401 actors, 2,502 directors, and 4,459 writers. The target nodes are movies, which are categorized into three classes.
    
\end{itemize}

\begin{table}[!htbp]
\centering
\caption{The Statistics Information of Datasets}
\label{tab:datasets}
\begin{tabular}{c|r|r|c}
\hline
Datasets                & Node-type: \#Nodes                    & Edge-type: \#Edges                                                                                 & Meta-paths                            \\ \hline
\multirow{3}{*}{ACM}    & Paper (P):\ \ \ \ \ \ \ \ 4,019         & \multirow{3}{*}{\begin{tabular}[c]{@{}r@{}}P-A:\ \ \ \ \ \  \ \  \ \  13,407 \\ P-S:\ \ \ \ \ \  \ \  \ \ \ \ 4,019\end{tabular}} & \multirow{3}{*}{\begin{tabular}[c]{@{}c@{}}PAP \\ PSP\end{tabular}}          \\ 
                        & Author (A):\ \ \ \ \ \ 7,167          &                                                                                                     &                                        \\ 
                        & Subject (S):\ \ \ \ \ \ \ \ \ \ 60        &                                                                                                     &                                        \\ \hline
\multirow{4}{*}{DBLP}   & Author (A):\ \ \ \  \ \ 4,057            & \multirow{4}{*}{\begin{tabular}[c]{@{}r@{}}P-A:\ \ \ \ \ \  \ \  \ \ 19,645 \\ P-T:\ \ \ \ \ \  \ \  \ \ 85,810 \\ P-C:\ \ \ \ \ \  \ \  \ \ 14,328\end{tabular}} & \multirow{4}{*}{\begin{tabular}[c]{@{}c@{}}APA \\ APTPA \\ APCPA\end{tabular}} \\ 
                        & Paper (P):\ \ \ \  \ \ 14,328            &                                                                                                     &                                        \\ 
                        & Term (T):\ \ \ \ \ \ \ \ \  7,723          &                                                                                                     &                                        \\ 
                        & Conference (C):\ \ \  \ \  20                &                                                                                                     &                                        \\ \hline
\multirow{3}{*}{AMiner} & Paper (P):\ \ \ \ \  \ \ \ 6,564           & \multirow{3}{*}{\begin{tabular}[c]{@{}r@{}}P-A:\ \ \ \ \ \  \ \  \ \ 18,007 \\ P-R:\ \ \ \ \ \  \ \  \ \ 58,831\end{tabular}}  & \multirow{3}{*}{\begin{tabular}[c]{@{}c@{}}PAP \\ PRP\end{tabular}}          \\ 
                        & Author (A):\ \ \ \ \ 13,329           &                                                                                                     &                                        \\ 
                        & Reference (R):\ \ 35,890              &                                                                                                     &                                        \\ \hline
\multirow{4}{*}{Freebase}   & Movie (M): \ \ \ \  \ \ 3,492            & \multirow{4}{*}{\begin{tabular}[c]{@{}r@{}}M-A: \ \ \ \ \ \ \  \ \ 65,341 \\ M-D:\ \ \ \ \ \  \ \ \  \ \ 3,762 \\ M-W:  \ \  \ \ \  \ \  \ \ 6,414\end{tabular}} & \multirow{4}{*}{\begin{tabular}[c]{@{}c@{}}MAM \\ MDM \\ MWM\end{tabular}} \\ 
                        & Actor (A):\ \ \ \ \  \ \ 33,401            &                                                                                                     &                                        \\ 
                        & Director (D):   \ \ \ \  2,502          &                                                                                                     &                                        \\ 
                        & Writer (W):\ \  \ \  \ \  4,459                &                                                                                                     &                                        \\ \hline
\end{tabular}
\end{table}

\textbf{Baselines.}
We compare the proposed ASHGCL with nine representative network representation learning methods. These methods include three unsupervised homogeneous methods (GraphSAGE~\cite{GragpSAGE}, GAE~\cite{GAE} and DGI~\cite{DGI}), three unsupervised heterogeneous methods (Mp2vec~\cite{Mp2vec}, HERec~\cite{HERec} and DMGI~\cite{DMGI}), two self-supervised heterogeneous methods (HeCo~\cite{HeCo} and MEOW~\cite{MEOW}) and a semi-supervised heterogeneous method HAN~\cite{HAN}.

\textbf{Implementation Details.}
For all methods, we perform 10 runs and generate 64-dimensional embeddings for evaluation, reporting the average results. For random walk-based methods (Mp2vec and HERec), we set the number of walks per node to 40, the walk length to 100, and the window size to 5. For homogeneous graph methods (GraphSAGE, GAE, DGI), we extract homogeneous subgraphs based on predefined meta-paths from the heterogeneous graph, apply these algorithms on each subgraph. For GraphSAGE, GAE, DGI, Mp2vec, and HERec, we conduct experiments using all relevant meta-paths and report the highest performance. For other configurations, we follow the settings in their original papers. Since many datasets only provide attributes for target node types, we use only these attributes for each dataset.  In cases where attributes for other node types are required, we generate one-hot vectors as their feature representations. We employ the following meta-paths for each dataset: for ACM, we use PAP (paper-author-paper) and PSP (paper-subject-paper); for DBLP, we use APA (author-paper-author), APCPA (author-paper-conference-paper-author) and APTPA (author-paper-term-paper-author); for Freebase, we use MAM (movie-actor-movie), MDM (movie-director-movie) and MWM (movie-writer-movie); for AMiner, we use PAP (paper-author-paper) and PRP (paper-reference-paper).

We implement ASHGCL in PyTorch, using Xavier initialization~\cite{Xavier} and Adam optimizer~\cite{Adam} to train the model. The code and dataset are available at \href{https://github.com/houlailyc/ASHGCL}{\texttt{https://github.com/houlailyc/ASHGCL}}. We fine-tune the learning rate from 5e-4 to 5e-3 and adjust early stopping patience between 5 and 50 epochs. We adjust \(\tau\) from 0.3 to 0.9 in increments of 0.1, and test dropout rates ranging from 0.2 to 0.5 with a step of 0.05. For the low-order structural view, we aggregate only the first-order heterogeneous neighbors of the target nodes. In the high-order structural view, we employ a single-layer GCN to aggregate neighbors within each meta-path. For the feature similarity view, we adjust from a one-layer to a four-layer GCN. We adjust the GCN architecture from one layer to four layers.

\textbf{Time Complexity Analysis.}
We analyze the time complexity of the ASHGCL framework, covering its four main components. In the \textbf{low-order structural view}, the first aggregation step assumes that each node has an average of \( K \) neighbors, with a complexity of \( O(|V| K d^2) \). This is followed by aggregation based on \( M \) types of neighbors, with a complexity of \( O(|V| M d^2) \). Thus, the overall time complexity of this part is \( O(|V| K d^2 + |V| M d^2) \). In the \textbf{high-order structural view}, the intra-metapath aggregation step assumes aggregation over \( L \) metapaths, with each metapath having an average of \( |P| \) edges, resulting in a complexity of \( O(L |P| d) \). This is followed by inter-metapath aggregation, with a complexity of \( O(|V| L d^2) \). Therefore, the overall time complexity of this part is \( O(L |P| d + |V| L d^2) \). In the \textbf{feature similarity view}, computing the cosine similarity between node pairs has a complexity of \( O(|V|^2 d) \), followed by GCN propagation with a complexity of \( O(|V| d^2) \). Thus, the overall time complexity of this part is \( O(|V|^2 d + |V| d^2) \). In the \textbf{cross-view contrastive learning}, the projection of embeddings from the three views has a complexity of \( O(|V| d) \). In the local contrastive learning step, each node has \( |\mathbb{P}_i| \) positive samples and \( |\mathbb{N}_i| \) negative samples, resulting in a complexity of \( O(|V| (|\mathbb{P}_i| + |\mathbb{N}_i|) d) \). The global contrastive learning step has a complexity of \( O(|V| d) \). Therefore, the overall complexity of this part is \( O(|V| d + |V| (|\mathbb{P}_i| + |\mathbb{N}_i|) d) \). Combining the above analysis, the overall computational complexity of ASHGCL is dominated by \( O(|V|^2 d) \).

\textbf{Evaluation Metrics.}
We assess the performance of node classification using three metrics: Macro-F1, Micro-F1, and AUC. Macro-F1 calculates the arithmetic mean of F1 scores for each class, ranging from 0 to 1. Micro-F1 computes a global F1 score by considering the total true positives, false negatives, and false positives across all classes, also ranging from 0 to 1. AUC measures the model's ability to distinguish between classes, particularly useful for imbalanced datasets.

For the node clustering task, we use normalized mutual information (NMI) and adjusted rand index (ARI) to assess the clustering results. NMI quantifies the normalized mutual information between the true partition and clustering results, ranging from 0 to 1. ARI is a corrected version of the Rand Index, measuring the similarity between the true partition and clustering assignments, ranging from -1 to 1.

\subsection{Node Classification}
\begin{table*}[!tb]
\centering
\caption{Quantitative results(\% ± $\sigma$) on node classification.}
\label{classification}
\begin{adjustbox}{width=\textwidth}
\begin{tabular}{lccccccccccc|c}
\toprule
Datasets & Metric & Split & GraphSAGE & GAE & Mp2vec & HERec & HAN & DGI & DMGI & HeCo &  MEOW & ASHGCL\\
\midrule

\multirow{9}{*}{ACM} & \multirow{3}{*}{Ma-F1} 
& 20 & 47.13±4.7 & 62.72±3.1 & 51.91±0.9 & 55.13±1.5 & 85.66±2.1 & 79.27±3.8 & 87.86±0.2 & 88.46±0.8  & 90.62±0.5 & \textbf{91.91±0.3}\\
& & 40 & 55.96±6.8 & 61.61±3.2 & 62.41±0.6 & 61.21±0.8 & 87.47±1.1 & 80.23±3.3 & 86.23±0.8 & 87.05±0.5  & 89.85±0.4 &\textbf{91.69±0.5} \\
& & 60 & 56.59±5.7 & 61.67±2.9 & 61.13±0.4 & 64.35±0.8 & 88.41±1.1 & 80.03±3.3 & 87.97±0.4 & 89.12±0.5  & 90.84±0.3 & \textbf{92.52±0.3}\\

\cmidrule{2-13}
& \multirow{3}{*}{Mi-F1} 
& 20 & 49.72±5.5 & 68.02±1.9 & 53.13±0.9 & 57.47±1.5 & 85.11±2.2 & 79.63±3.5 & 87.60±0.8 & 88.04±0.8 & 90.63±0.3 & \textbf{91.52±0.3} \\
& & 40 & 60.98±3.5 & 66.38±1.9 & 64.43±0.6 & 62.62±0.9 & 87.21±1.2 & 80.41±3.0 & 86.02±0.9 & 87.34±0.5  & 90.14±0.4 & \textbf{91.42±0.3} \\
& & 60 & 60.72±4.3 & 65.71±2.2 & 62.72±0.3 & 65.15±0.9 & 88.10±1.2 & 80.15±3.2 & 87.82±0.5 & 88.31±0.5  & 90.76±0.3 & \textbf{92.36±0.2}\\

\cmidrule{2-13}
& \multirow{3}{*}{AUC} 
& 20 & 65.88±3.7 & 79.50±2.4 & 71.66±0.7 & 75.44±1.3 & 93.47±1.5 & 91.47±2.3 & 96.72±0.3 & 96.25±0.3 &  97.60±0.2 & \textbf{98.15±0.2}\\
& & 40 & 71.06±5.2 & 79.14±2.5 & 80.48±0.4 & 79.84±0.5 & 94.84±0.9 & 91.52±2.3 & 96.35±0.3 & 96.27±0.4 &  97.74±0.1 & \textbf{98.37±0.3} \\
& & 60 & 70.45±6.2 & 77.90±2.8 & 79.33±0.4 & 81.64±0.7 & 94.68±1.4 & 91.41±1.9 & 96.79±0.2 &96.12±0.3  & 98.05±0.1 & \textbf{98.25±0.2}\\
\midrule

\multirow{9}{*}{DBLP} & \multirow{3}{*}{Ma-F1} 
& 20 & 71.97±8.4 & 90.90±0.1 & 88.98±0.2 & 89.57±0.4 & 89.31±0.9 & 87.93±2.4 & 89.94±0.4 & 90.69±0.2 &  92.16±0.5 & \textbf{92.94±0.4}\\
& & 40 & 73.69±8.4 & 89.60±0.3 & 88.68±0.2 & 89.73±0.4 & 88.87±1.0 & 88.62±0.6 & 89.25±0.4 & 90.17±0.3  & 91.15±0.3 & \textbf{92.30±0.4}\\
& & 60 & 73.86±8.1 & 90.08±0.2 & 90.25±0.1 & 90.18±0.3 & 89.20±0.8 & 89.19±0.9 & 89.46±0.6 & 90.52±0.3  & 93.05±0.5 & \textbf{93.29±0.5}\\
\cmidrule{2-13}
& \multirow{3}{*}{Mi-F1} 
& 20 & 71.44±8.7 & 91.55±0.1 & 89.67±0.1 & 90.24±0.4 & 90.16±0.9 & 88.72±2.6 & 90.78±0.3 & 91.40±0.2  & 92.64±0.4 & \textbf{93.27±0.6}\\
& & 40 & 73.61±8.6 & 90.00±0.3 & 89.14±0.2 & 90.15±0.4 & 89.47±0.9 & 89.22±0.5 & 89.92±0.4 & 90.54±0.3  & 91.51±0.3 & \textbf{92.65±0.3}\\
& & 60 & 74.05±8.3 & 90.95±0.2 & 91.17±0.1 & 91.01±0.3 & 90.34±0.8 & 90.35±0.8 & 90.66±0.5 & 91.54±0.2  & 93.50±0.5 & \textbf{93.97±0.3}\\
\cmidrule{2-13}
& \multirow{3}{*}{AUC} 
& 20 & 90.59±4.3 & 98.15±0.1 & 97.69±0.0 & 98.21±0.2 & 98.07±0.6 & 96.99±1.4 & 97.75±0.3 & 98.08±0.1 &  98.87±0.2 & \textbf{98.92±0.2}\\
& & 40 & 91.42±4.0 & 97.85±0.1 & 97.08±0.0 & 97.93±0.1 & 97.48±0.6 & 97.12±0.4 & 97.23±0.2 & 98.06±0.1  & 98.55±0.2 & \textbf{98.66±0.3}\\
& & 60 & 91.73±3.8 & 98.37±0.1 & 98.00±0.0 & 98.49±0.1 & 97.96±0.5 & 97.76±0.5 & 97.72±0.4 & 98.68±0.1  & 99.05±0.1 & \textbf{99.10±0.2}\\
\midrule

\multirow{9}{*}{AMiner} & \multirow{3}{*}{Ma-F1} 
& 20 & 42.46±2.5 & 60.22±2.0 & 54.78±0.5 & 58.32±1.1 & 56.07±3.2 & 51.61±3.2 & 59.50±2.1 & 68.62±1.1 & 69.79±0.9 & \textbf{74.80±1.8}\\
& & 40 & 45.77±1.5 & 65.66±1.5 & 64.77±0.5 & 64.50±0.7 & 63.85±1.5 & 54.72±2.6 & 61.92±2.1 & 71.91±0.5 & 70.26±1.0 & \textbf{76.13±1.1}\\
& & 60 & 44.91±2.0 & 63.74±1.6 & 60.65±0.3 & 65.53±0.7 & 62.02±1.2 & 55.45±2.4 & 61.15±2.5 & 75.45±1.8  & 72.79±1.0 & \textbf{76.28±1.2}\\
\cmidrule{2-13}
& \multirow{3}{*}{Mi-F1} 
& 20 & 49.68±3.1 & 65.78±2.9 & 60.82±0.4 & 63.64±1.1 & 68.86±4.6 & 62.39±3.9 & 63.93±3.3 & 75.06±1.3  & 76.26±0.5 & \textbf{82.40±0.6}\\
& & 40 & 52.10±2.2 & 71.34±1.8 & 69.66±0.6 & 71.57±0.7 & 76.89±1.6 & 63.87±2.9 & 63.60±2.5 & 79.56±0.7 &  75.90±0.6 & \textbf{83.07±0.9}\\
& & 60 & 51.36±2.2 & 67.70±1.9 & 63.92±0.5 & 69.76±0.8 & 74.73±1.4 & 63.10±3.0 & 62.51±2.6 & 81.94±1.4  & 77.92±0.4 & \textbf{83.15±0.7}\\
\cmidrule{2-13}
& \multirow{3}{*}{AUC} 
& 20 & 70.86±2.5 & 85.39±1.0 & 81.22±0.3 & 83.35±0.5 & 78.92±2.3 & 75.89±2.2 & 85.34±0.9 & 89.74±0.6  & 90.45±0.2 & \textbf{90.55±0.5}\\
& & 40 & 74.44±1.3 & 88.29±1.0 & 88.82±0.2 & 88.70±0.4 & 80.72±2.1 & 77.86±2.1 & 88.02±1.3 & 92.26±0.6  & 92.43±0.3 & \textbf{93.75±0.4}\\
& & 60 & 74.16±1.3 & 86.92±0.8 & 85.57±0.2 & 87.74±0.5 & 80.39±1.5 & 77.21±1.4 & 86.20±1.7 & 92.38±0.7  & 92.64±0.2 & \textbf{92.70±0.5}\\
\midrule

\multirow{9}{*}{Freebase} & \multirow{3}{*}{Ma-F1} 
& 20 & 45.14±4.5 & 53.81±0.6 & 53.96±0.7 & 55.78±0.5 & 53.16±2.8 & 54.90±0.7 & 55.79±0.9 & 59.52±0.5 & 55.01±0.9 &\textbf{63.02±0.5} \\

& & 40 & 44.88±4.1 & 52.44±2.3 & 57.80±1.1 & 59.28±0.6 & 59.63±2.3 & 53.40±1.4 & 49.88±1.9 & 61.15±0.4 & 55.19±0.8 &\textbf{62.68±0.3} \\

& & 60 & 45.16±3.1 & 50.65±0.4 & 55.94±0.7 & 56.50±0.4 & 56.77±1.7 & 53.81±1.1 & 52.10±0.7 & 58.85±1.0 & 57.46±0.5 &\textbf{60.95±0.6} \\
\cmidrule{2-13}
& \multirow{3}{*}{Mi-F1} 
& 20 & 54.83±3.0 & 55.20±0.7 & 56.23±0.8 & 57.92±0.5 & 57.24±3.2 & 58.16±0.9 & 58.26±0.9 & 63.56±0.3 &58.86±0.7 &\textbf{67.01±0.3} \\
& & 40 & 57.08±3.2 & 56.05±2.0 & 61.01±1.3 & 62.71±0.7 & 63.74±2.7 & 57.82±0.8 & 54.28±1.6 & 64.12±0.5 & 59.64±0.5 &\textbf{66.22±0.5} \\
& & 60 & 55.92±3.2 & 53.85±0.4 & 58.74±0.8 & 58.57±0.5 & 61.06±2.0 & 57.96±0.7 & 56.69±1.2 & 62.04±0.9 & 59.21±0.4 &\textbf{64.80±0.3} \\
\cmidrule{2-13}
& \multirow{3}{*}{AUC} 
& 20 & 67.63±5.0 & 73.03±0.7 & 71.78±0.7 & 73.89±0.4 & 73.26±2.1 & 72.80±0.6 & 73.19±1.2 & 76.03±0.5 & 73.51±0.7 & \textbf{77.29±0.5}\\
& & 40 & 66.42±4.7 & 74.05±0.9 & 75.51±0.8 & 76.08±0.4 & 77.74±1.2 & 72.97±1.1 & 70.77±1.6 & 78.10±0.3 & 75.32±0.8 &\textbf{78.35±0.4} \\
& & 60 & 66.78±3.5 & 71.75±0.4 & 74.78±0.4 & 74.89±0.4 & 75.69±1.5 & 73.32±0.9 & 73.17±1.4 & 78.15±0.3 & 75.44±0.7 &\textbf{78.19±0.4} \\
\bottomrule
\end{tabular}
\end{adjustbox}
\end{table*}

To evaluate the effectiveness of our ASHGCL model in node classification tasks, we use the learned node embeddings to train a linear classifier. We select 20, 40, 60 labeled nodes per class as training sets, with 1,000 nodes each for validation and testing. We employ Macro-F1, Micro-F1, and AUC as evaluation metrics, where higher values indicate better performance. The results are presented in Table~\ref{classification}, where the best performance for each metric is highlighted in bold. As shown, ASHGCL consistently outperforms baseline methods across all datasets and training set sizes, even surpassing the semi-supervised method HAN. Notably, ASHGCL achieves superior performance compared to state-of-the-art models like HeCo and MEOW. For example, on the AMiner dataset with the 40\% training ratio, MEOW, a state-of-the-art method, achieves Macro-F1 and Micro-F1 scores of 70.26\% and 75.90\%, respectively. In contrast, our proposed ASHGCL achieves 76.13\% and 83.07\%. These significant improvements demonstrate the effectiveness of our approach in learning discriminative node representations for heterogeneous graphs.

\subsection{Node Clustering}

\begin{figure}[!tb]
    \centering
    \subcaptionbox{HeCo\label{fig:scatter1}}
    [0.4\textwidth]{\includegraphics[width=\linewidth]{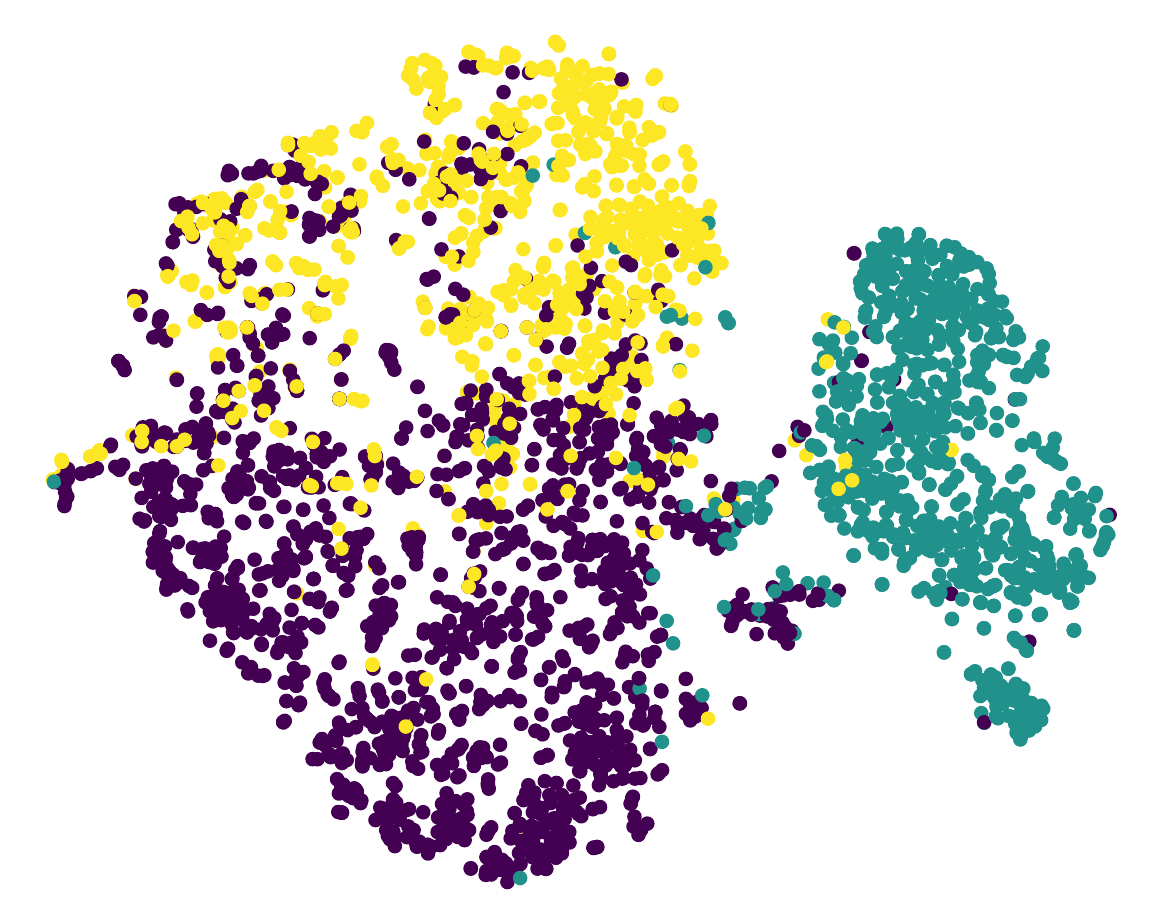}}
    \hfill
    \subcaptionbox{MEOW\label{fig:scatter2}}
    [0.4\textwidth]{\includegraphics[width=\linewidth]{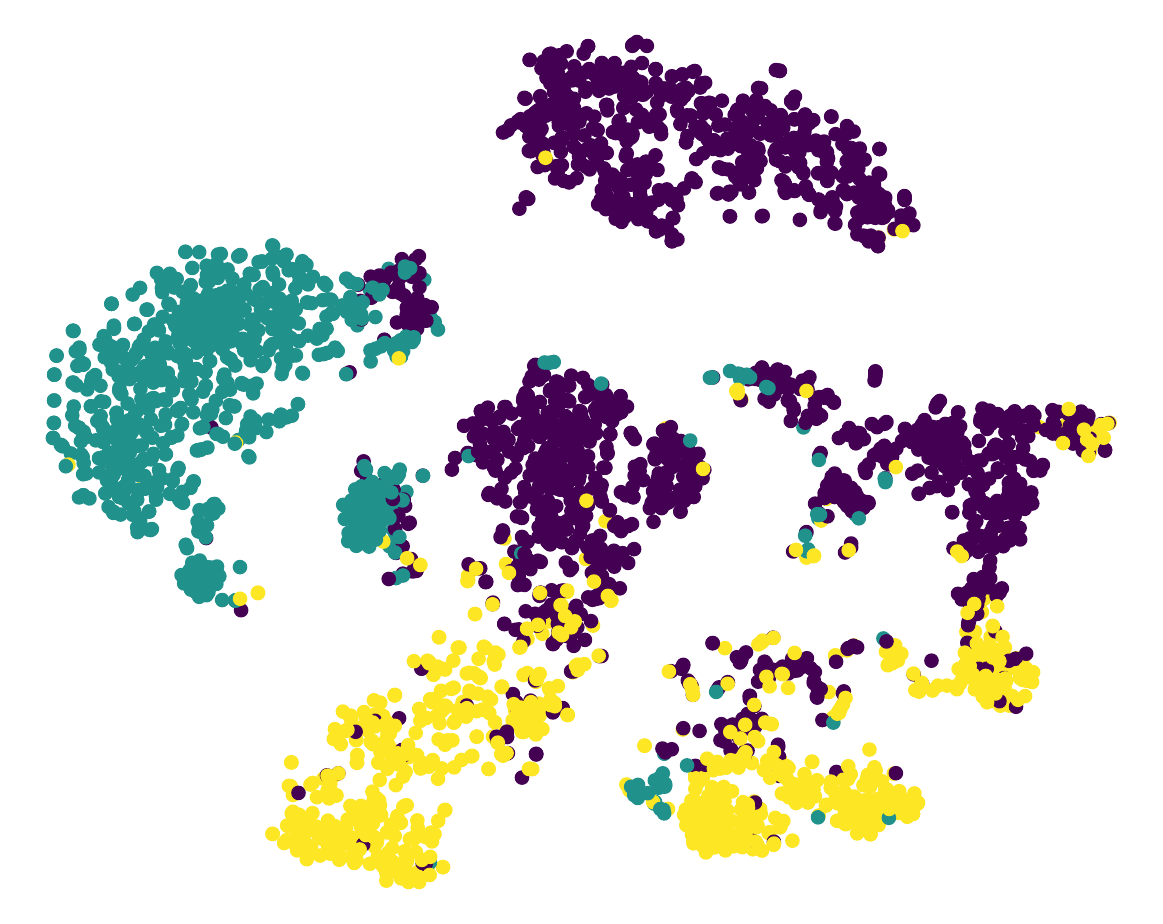}}
    
    \vspace{1em}
    
    \subcaptionbox{DMGI\label{fig:scatter3}}
    [0.4\textwidth]{\includegraphics[width=\linewidth]{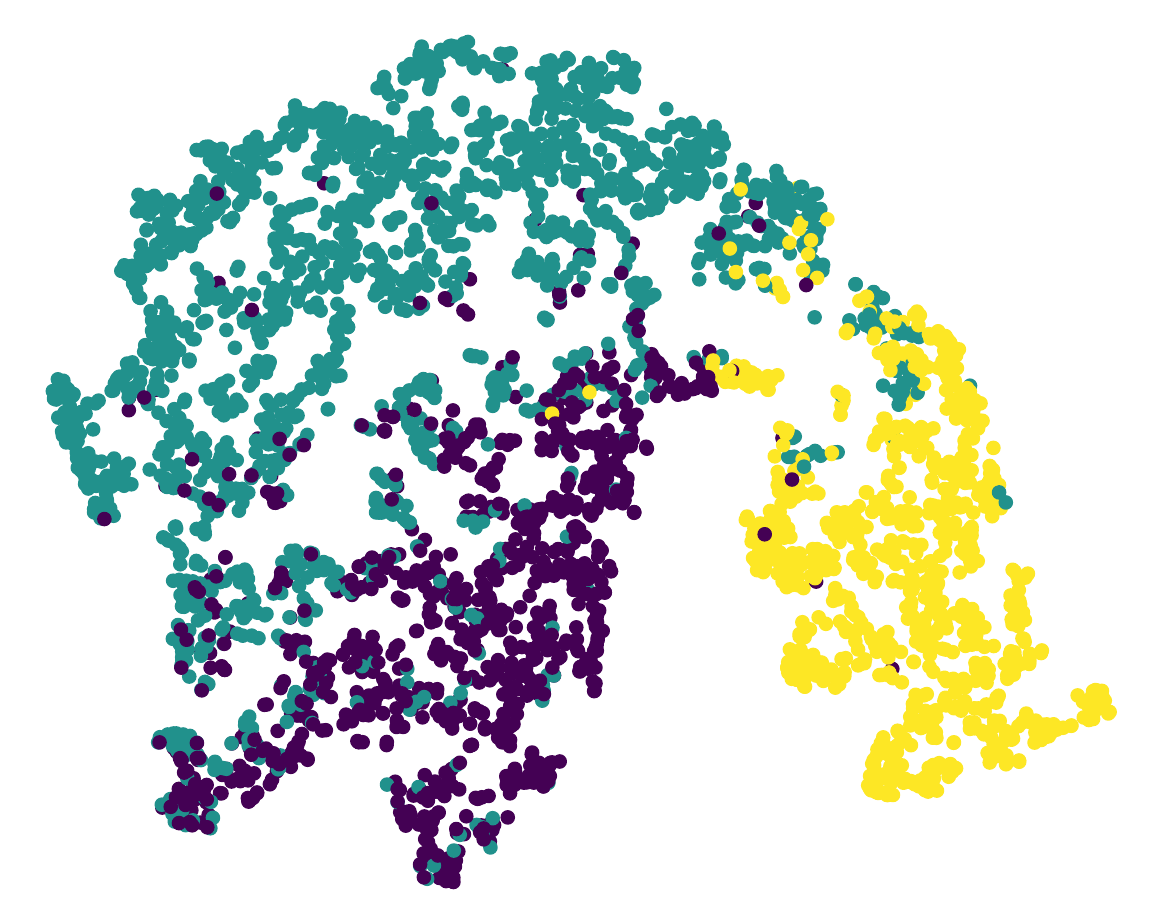}}
    \hfill
    \subcaptionbox{ASHGCL\label{fig:scatter4}}
    [0.4\textwidth]{\includegraphics[width=\linewidth]{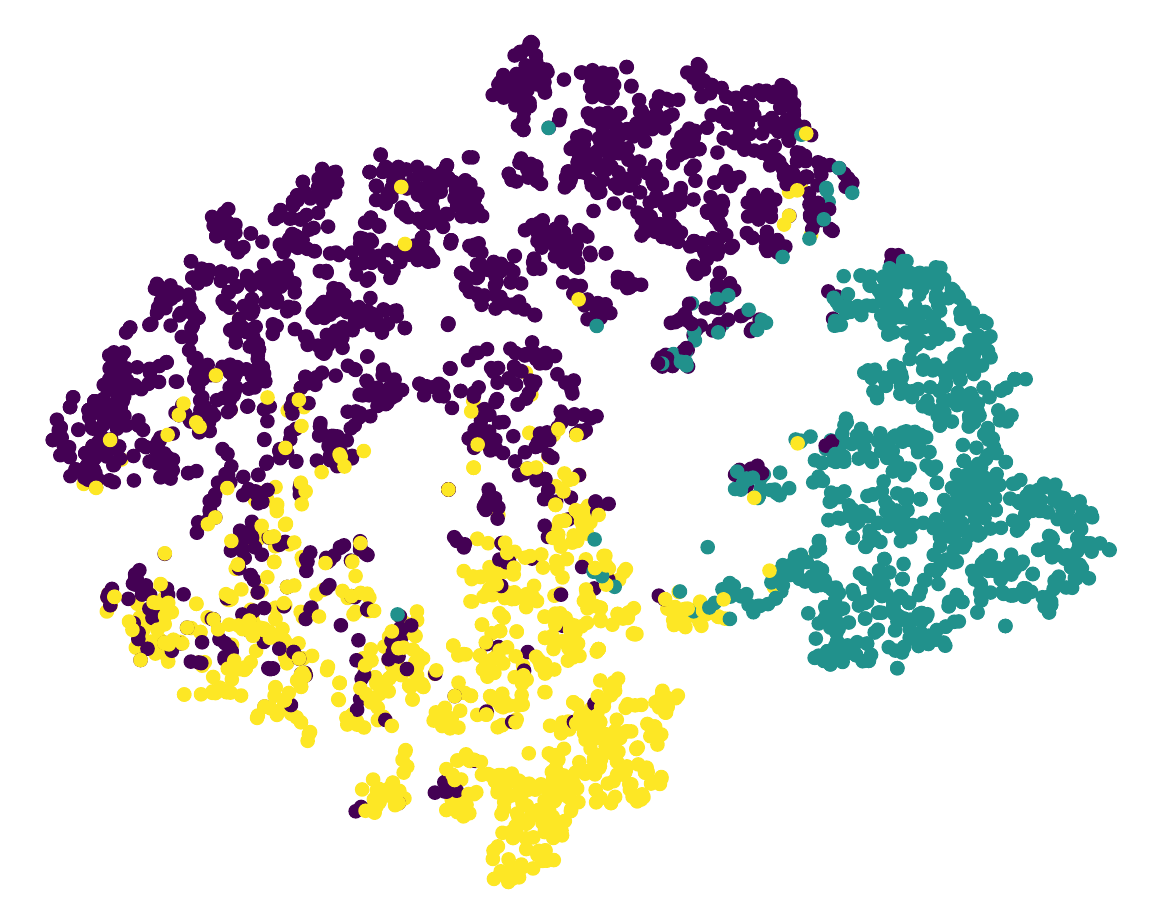}}
    
    \caption{Visualization of the embeddings learned by ASHGCL and baselines on ACM dataset. The Silhouette scores for (a) (b) (c) (d) are 0.2403, 0.2758, 0.3355 and 0.3864, respectively.}
    \label{visual}
\end{figure}

In this section, we analyze the node clustering performance of our ASHGCL model. We utilize the K-means algorithm~\cite{kmeans} on the learned embeddings to perform clustering tasks. We employ NMI and ARI as evaluation metrics for clustering quality, with higher values indicating better results. To ensure the reliability of our results, we repeat each experiment 10 times and report the average scores. The clustering results are presented in Table~\ref{clustering}, with the best performance for each metric highlighted in bold. As demonstrated, ASHGCL consistently achieves superior performance across all datasets compared to the baseline methods. Particularly noteworthy are the results on the ACM and AMiner datasets. In ACM, ASHGCL surpasses the next-best method (HeCo) by 9.28 and 13.51 percentage points in NMI and ARI, respectively. The performance gap is even more pronounced on the AMiner dataset, where ASHGCL outperforms the second-best methods by 8.99 and 21.35 percentage points in NMI and ARI respectively. These substantial improvements clearly demonstrate the superior performance of our proposed method.

\begin{table}[!htbp]
\centering
\caption{Quantitative results(\%) on node clustering.}
\label{clustering}
\resizebox{\columnwidth}{!}{
\begin{tabular}{c|cc|cc|cc|cc}
\hline
Datasets  & \multicolumn{2}{c|}{ACM} & \multicolumn{2}{c|}{DBLP} & \multicolumn{2}{c|}{AMiner}& \multicolumn{2}{c}{Freebase} \\
\hline
 Metrics & NMI & ARI & NMI & ARI & NMI & ARI& NMI & ARI \\
\hline
GraphSage & 29.20 & 27.72 & 51.50 & 36.40 & 15.74 & 10.10&9.05 &10.49 \\
GAE &  27.42 & 24.49 & 72.59 & 77.31 & 28.58 & 20.90&19.03 &14.10 \\
Mp2vec  & 48.43 & 34.65 & 73.55 & 77.70 & 30.80 & 25.26&16.47& 17.32\\
HERec  & 47.54 & 35.67 & 70.21 & 73.99 & 27.82 & 20.16 &19.76&19.36\\
DGI  & 51.73 & 41.16 & 59.23 & 61.85 & 22.06 & 15.93 &18.34&11.29\\
DMGI  & 51.66 & 46.64 & 70.06 & 75.46 & 19.24 & 20.09 &16.38&16.91\\
HeCo  & 59.26 & 58.36 & 71.58 & 76.93 & 29.02 & 18.72 &18.25&18.42\\
MEOW  & 51.12 & 45.20 & 73.78 & 79.58 & 24.05 & 17.66&13.25&14.40 \\
\hline
ASHGCL  & \textbf{68.54}& \textbf{71.87} &\textbf{77.47} & \textbf{82.46} &\textbf{39.79} & \textbf{47.95}&\textbf{21.35}&\textbf{22.02}\\
\hline
\end{tabular}
}
\end{table}

\subsection{Visualization}
To provide an intuitive demonstration of performance, we visualized the learned embeddings on the ACM dataset. Specifically, we used t-SNE~\cite{sne} to visualize the representations learned from HeCo, MOEW, DMGI, and ASHGCL, with the results presented in Figure~\ref{visual}.

As we can see, HeCo and MEOW demonstrate some separation between node types but with substantial overlap.  DMGI presents clearer boundaries between different node types, though some mixing is still evident. Our proposed ASHGCL method outperforms the baselines, exhibiting the most distinct separation of node types and forming more compact clusters, demonstrating the effectiveness of our approach.

\subsection{Ablation Study}

In this section, we investigate the impact of key components within our model on overall performance through ablation experiments. We conducted ablation studies on the proposed view configuration, node sampling strategy and contrast configuration separately. We perform node classification tasks on the ACM and DBLP datasets with 40\% labeled data.

\textbf{View Configuration.}
Existing methods often capture the high-order structural information of graphs through meta-paths to construct views, which may overlook the low-order structural information and the inherent information of features. We design two variants, namely ASHGCL\_lo and ASHGCL\_fe, to explore the impact of these views on performance. In ASHGCL\_lo, the low-order structural view is omitted, while in ASHGCL\_fe, the feature similarity view is excluded.

\begin{figure}[!tb]
    \centering
    \begin{subfigure}[b]{0.4\textwidth}
        \centering
        \includegraphics[width=\textwidth]{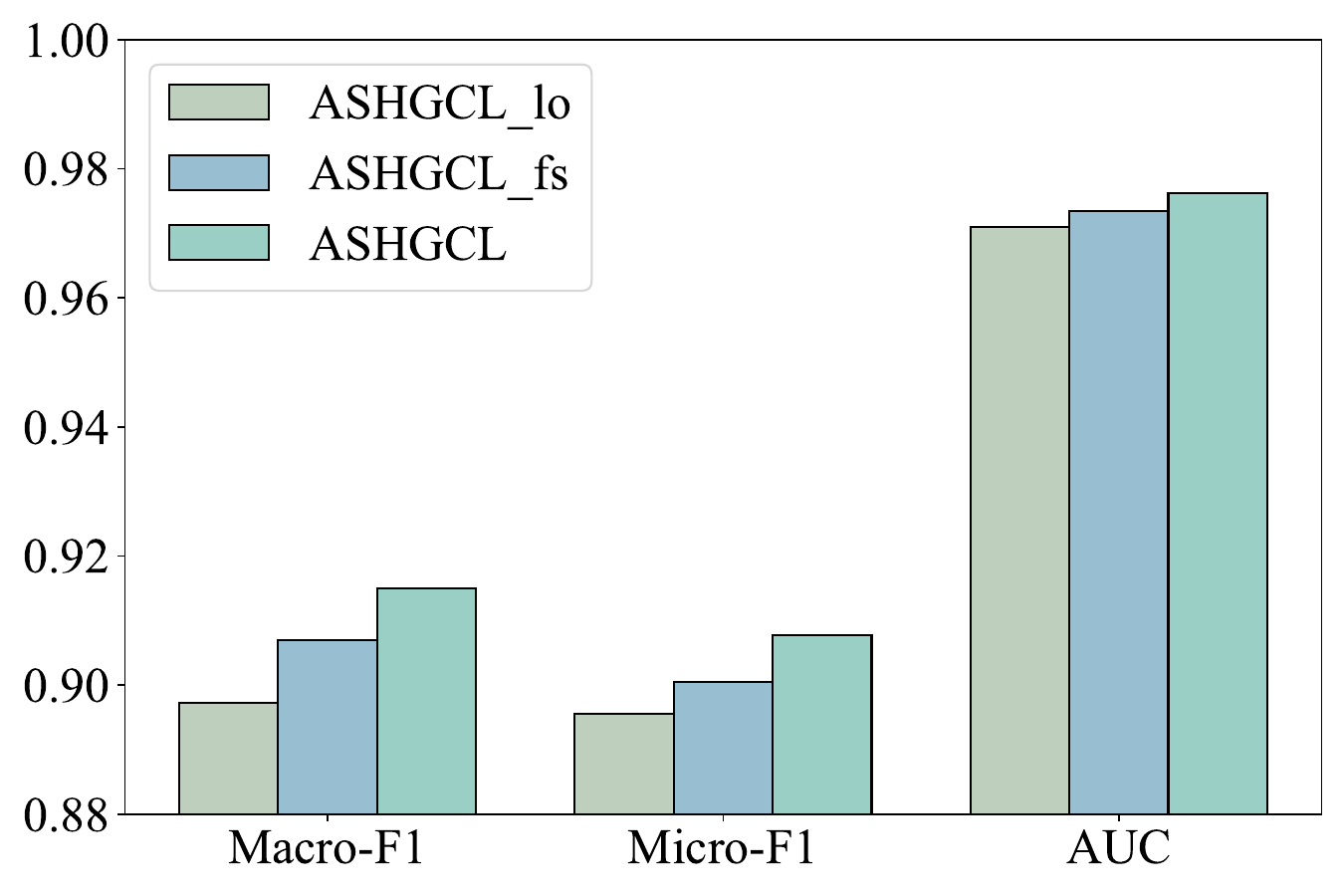}
        \caption{ACM}
        \label{abalationview_acm}
    \end{subfigure}
    \begin{subfigure}[b]{0.4\textwidth}
        \centering
        \includegraphics[width=\textwidth]{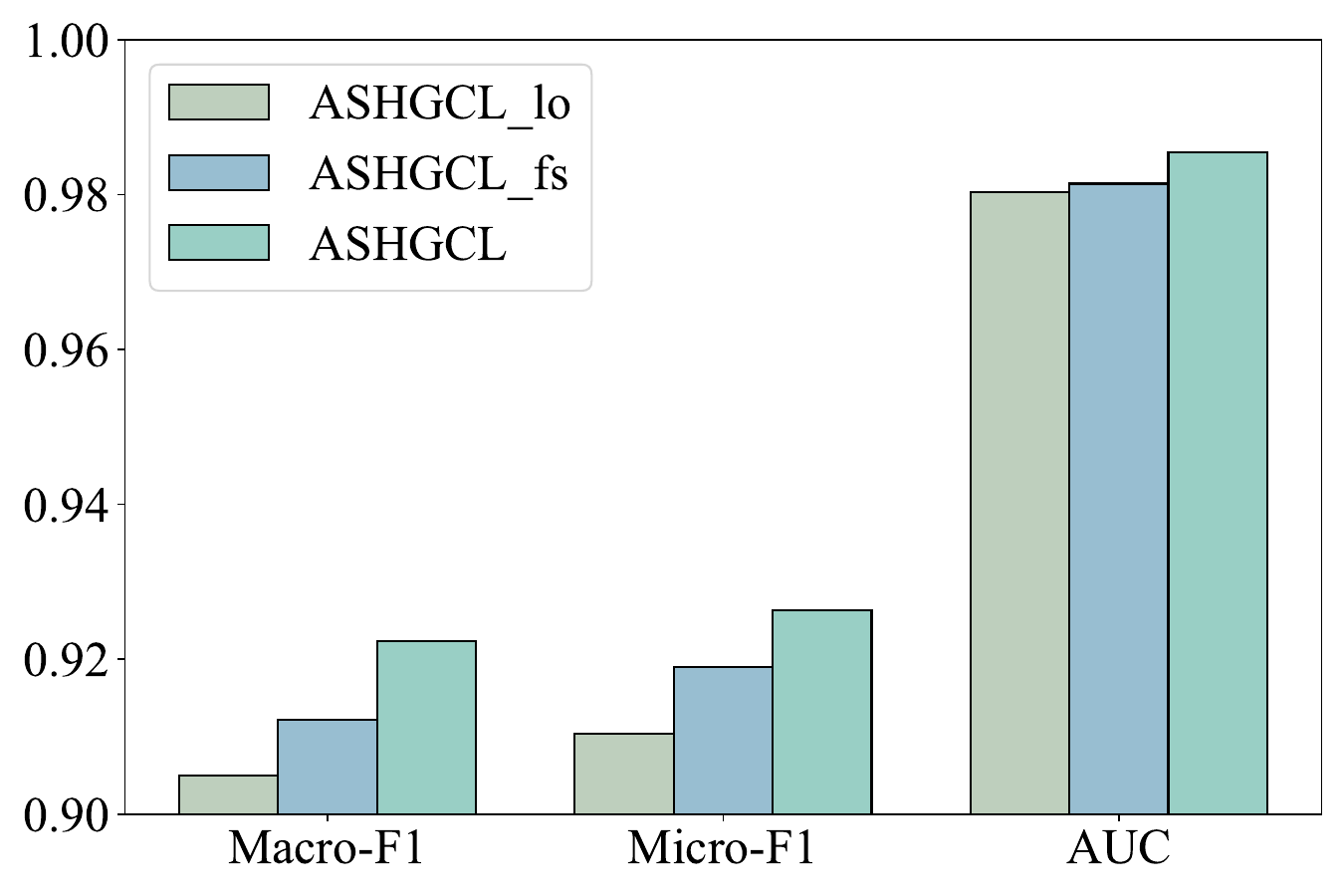}
        \caption{DBLP}
        \label{abalationview_dblp}
    \end{subfigure}
    \caption{Ablation experimental results of the view configuration.}
    \label{abalationview}
\end{figure}

The results of ASHGCL and the two variants are shown in Figure~\ref{abalationview}. It is evident that ASHGCL consistently outperforms both variants, demonstrating the effectiveness of our proposed three-view strategy. Additionally, we observe that ASHGCL\_fe consistently surpasses ASHGCL\_lo, indicating that the low-order structural information of nodes is more critical than the feature information.

\textbf{Sampling Strategy.}
To demonstrate the effectiveness of our proposed sampling strategy, we design two variants, ASHGCL\_tp and ASHGCL\_sm. In ASHGCL\_tp, only structural-aware positive samples are used for training, whereas in ASHGCL\_sm, only attribute-aware positive samples are utilized for training.

\begin{figure}[!tb]
    \centering
    \begin{subfigure}[b]{0.4\textwidth}
        \centering
        \includegraphics[width=\textwidth]{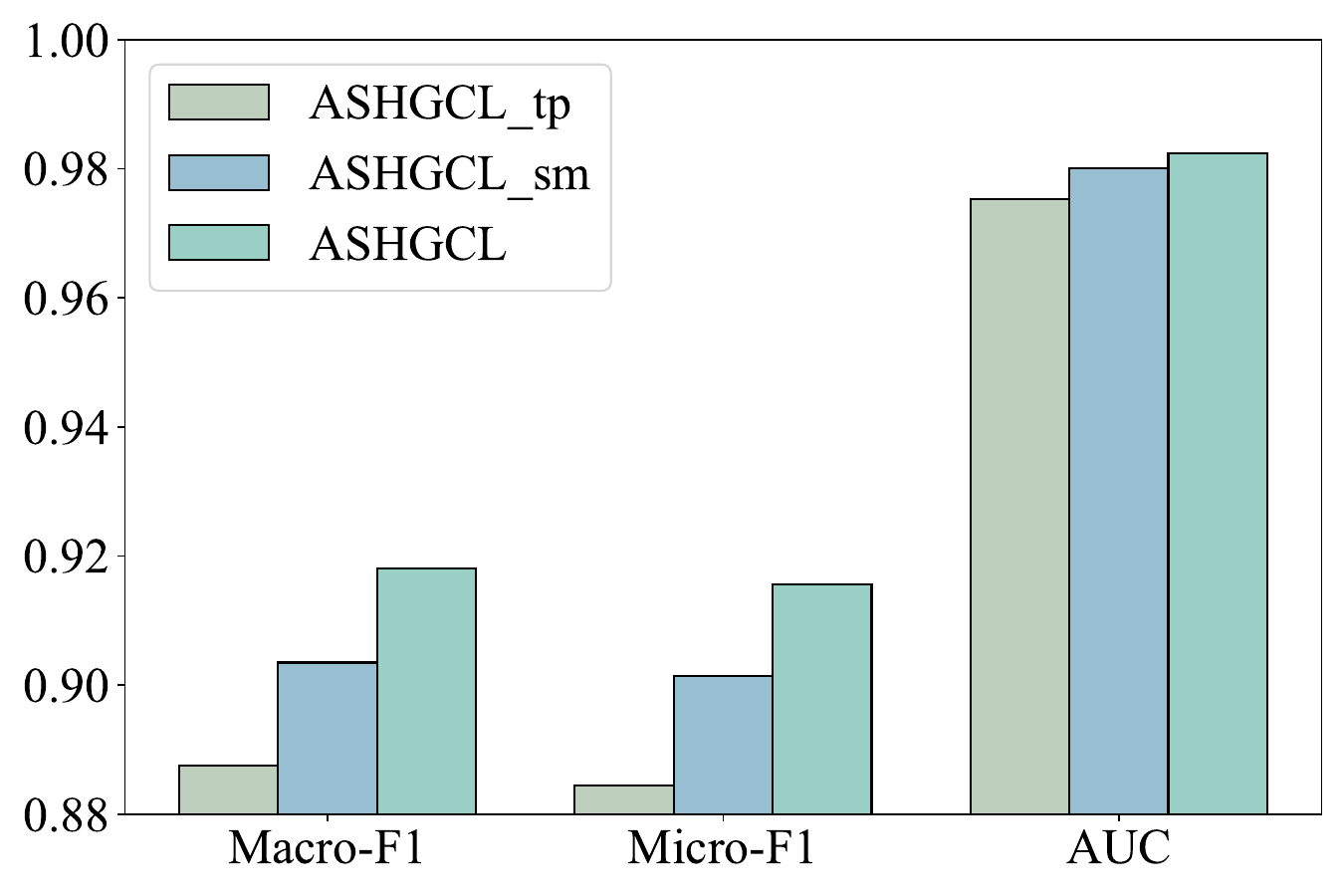}
        \caption{ACM}
        \label{abalationpos_acm}
    \end{subfigure}
    \begin{subfigure}[b]{0.4\textwidth}
        \centering
        \includegraphics[width=\textwidth]{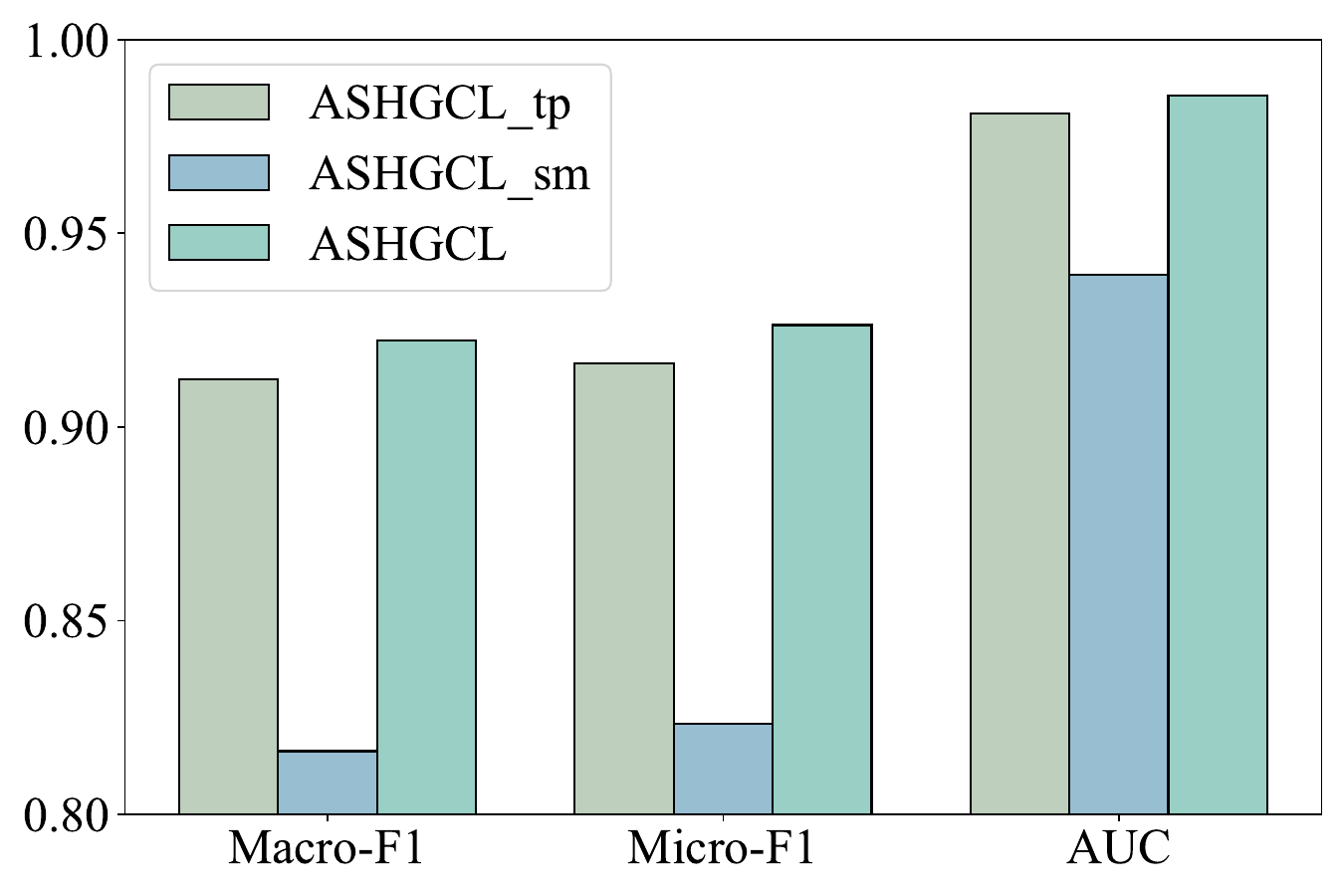}
        \caption{DBLP}
        \label{abalationpos_dblp}
    \end{subfigure}
    \caption{Ablation experimental results of the sampling strategy.}
    \label{abalationpos}
\end{figure}

The results of ASHGCL and the two variants are shown in Figure~\ref{abalationpos}. It is evident that ASHGCL consistently outperforms the other two variants, indicating the need to consider both topological structure and node attributes during sampling. It can also be observed that on the DBLP dataset, the performance of ASHGCL\_sm is significantly lower than that of the other two models. This is because the node attributes in the DBLP dataset are sparse, with many dimensions having zero values, leading to insufficient node attribute information. Relying solely on this limited node attribute information hinders the model's ability to train effectively. Therefore, we believe that if higher-quality node attribute information was available, the performance of our model could be further enhanced.

\textbf{Contrast Configuration.}
To demonstrate the effectiveness of the global contrast strategy, we design a variant, ASHGCL\_g. In ASHGCL\_g, only local contrastive learning is employed, while the global contrastive learning component is omitted.

\begin{figure}[!tb]
    \centering
    \begin{subfigure}[b]{0.4\textwidth}
        \centering
        \includegraphics[width=\textwidth]{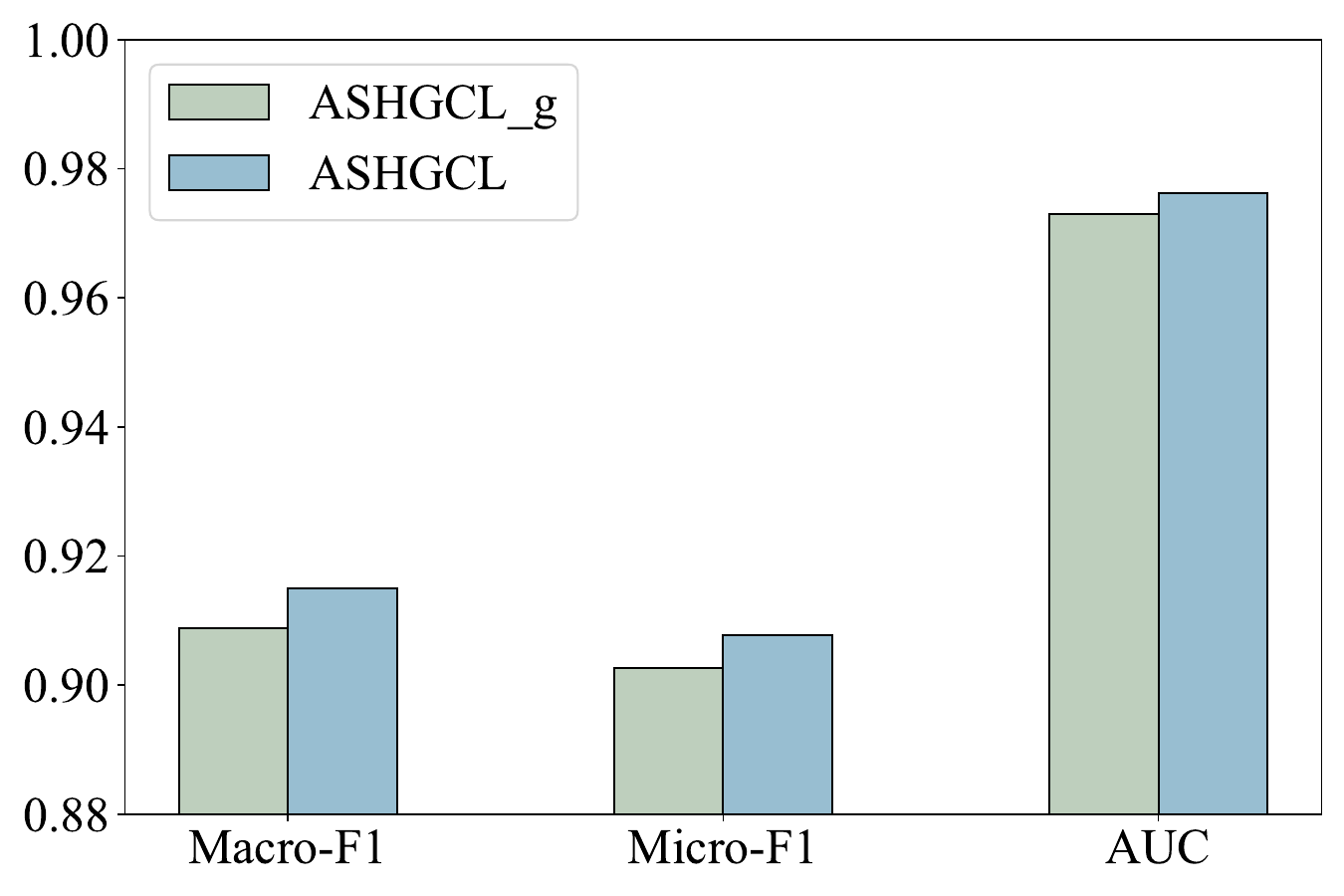}
        \caption{ACM}
        \label{abalationglobal_acm}
    \end{subfigure}
    \begin{subfigure}[b]{0.4\textwidth}
        \centering
        \includegraphics[width=\textwidth]{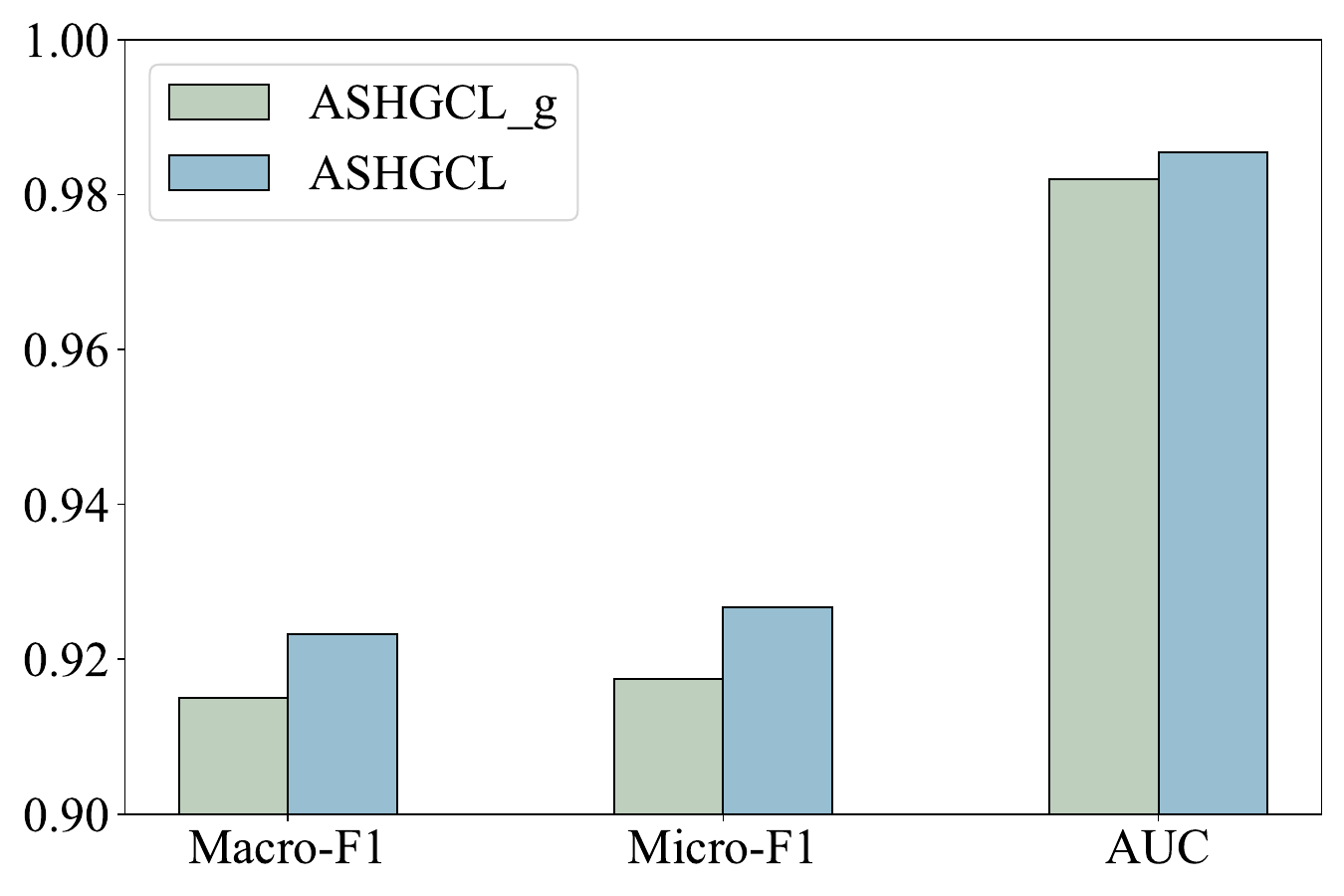}
        \caption{DBLP}
        \label{abalationglobal_dblp}
    \end{subfigure}
    \caption{Ablation experimental results of the contrast configuration.}
    \label{abalationglobal}
\end{figure}

The results of ASHGCL and the variant are shown in Figure~\ref{abalationglobal}. It is evident that ASHGCL consistently outperforms ASHGCL\_g, demonstrating the efficacy of global contrastive learning. While existing methods typically employ only local contrastive learning to capture node-level information, our approach incorporates global contrastive learning to capture coarse-grained graph-level information. Although utilizing local contrastive learning alone can yield satisfactory results, the inclusion of global contrastive learning as a complementary component enables the model to acquire additional graph-level knowledge. This enhancement leads to an improvement in model performance, with an increase ranging from 0.5\% to 1\%.

\subsection{Hyperparameter Analysis}

In this section, we discuss the impact of three key hyperparameters in our proposed method, ASHGCL: the temperature hyperparameter \(\tau\), the number of positive samples \(k\) and the feature similarity threshold $\theta$. We conduct node classification experiments on the ACM and DBLP datasets to evaluate how these parameters influence model performance, using Micro-F1, Macro-F1, and AUC scores as metrics. 

\begin{figure}[!tb]
    \centering
    \begin{subfigure}[b]{0.43\textwidth}
        \centering
        \includegraphics[width=\textwidth]{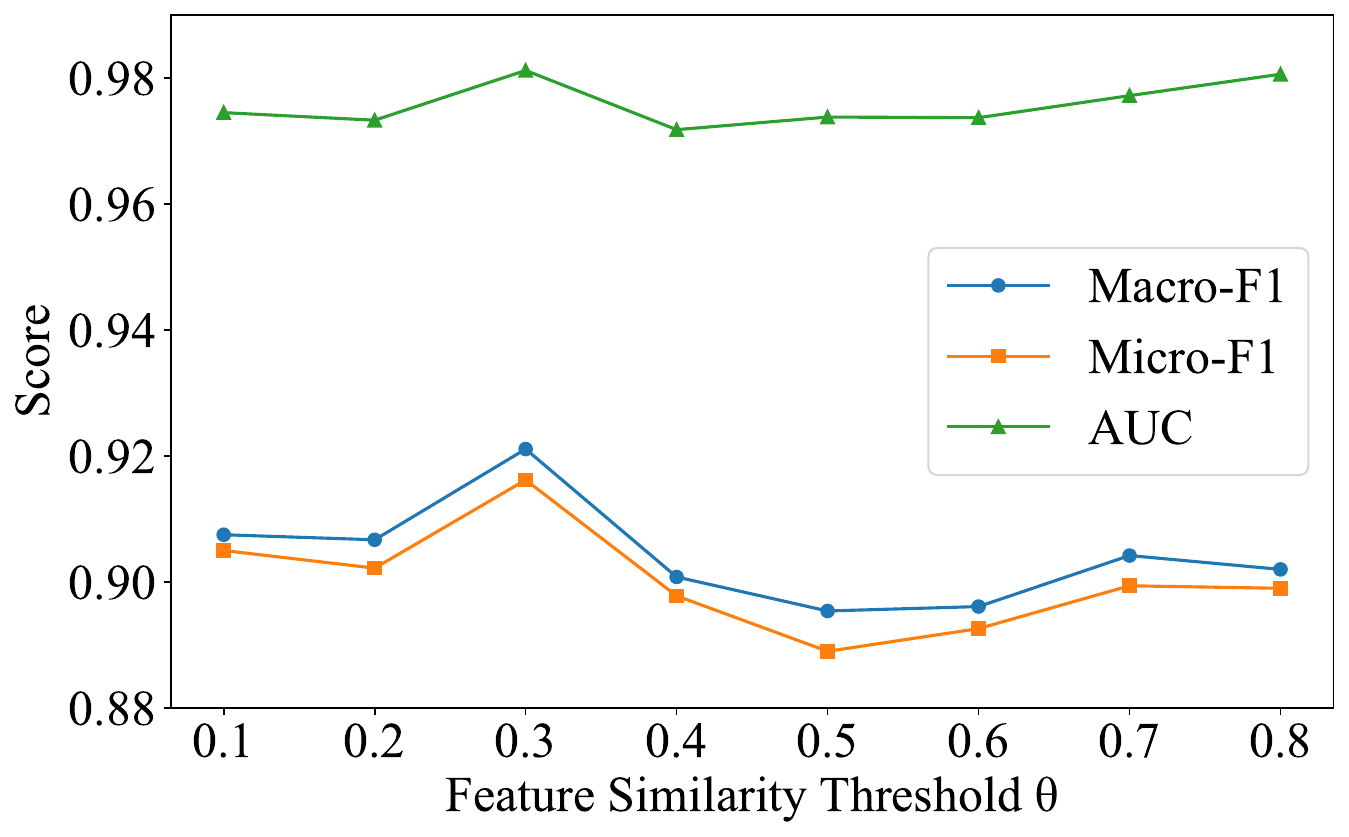}
        \caption{ACM}
        \label{feparam_acm}
    \end{subfigure}
    \begin{subfigure}[b]{0.43\textwidth}
        \centering
        \includegraphics[width=\textwidth]{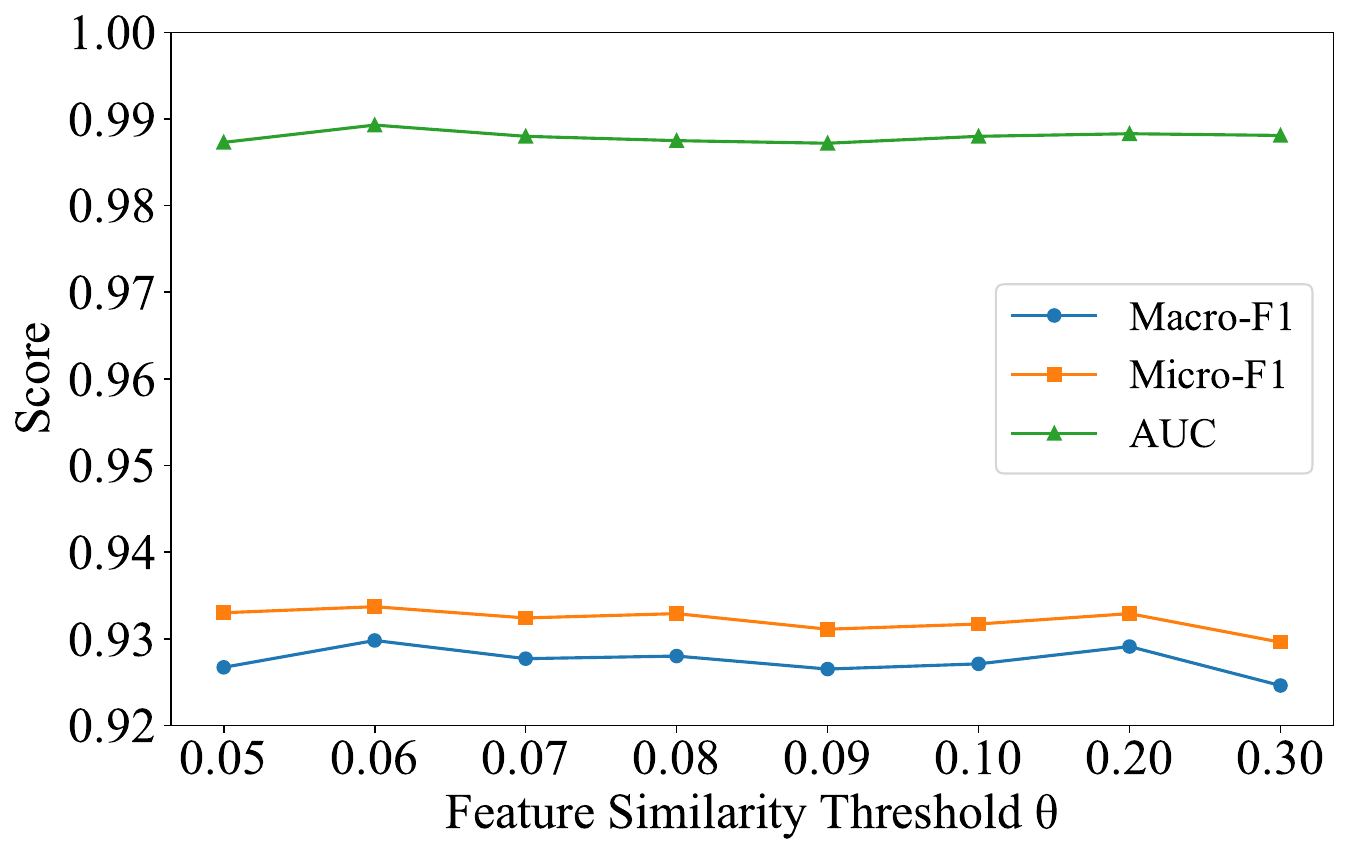}
        \caption{DBLP}
        \label{feparam_dblp}
    \end{subfigure}
    \caption{Analysis of the feature similarity threshold $\theta$ impact on model performance.}
    \label{feparam}
\end{figure}

\textbf{Analysis of $\theta$.}
The hyperparameter $\theta$ determines the threshold for feature similarity when constructing the feature similarity view. As shown in Figure~\ref{feparam}, for the ACM dataset, the performance of the model improves consistently as $\theta$ increases, reaching its peak at $\theta=0.3$, after which it begins to decline, with the lowest performance observed at $\theta=0.5$. This trend suggests that an appropriate threshold is crucial for balancing the inclusion of relevant nodes and the exclusion of noise. In contrast, the DBLP dataset exhibits its optimal performance at a much lower threshold of $\theta=0.06$, with relatively minor fluctuations in performance as $\theta$ varies. This behavior can be attributed to the significant attribute sparsity in the DBLP dataset, resulting in generally low feature similarities between nodes. Consequently, adjusting the similarity threshold introduces fewer new nodes to the graph, leading to a less pronounced impact on model performance.

\begin{figure}[!tb]
    \centering
    \begin{subfigure}[b]{0.43\textwidth}
        \centering
        \includegraphics[width=\textwidth]{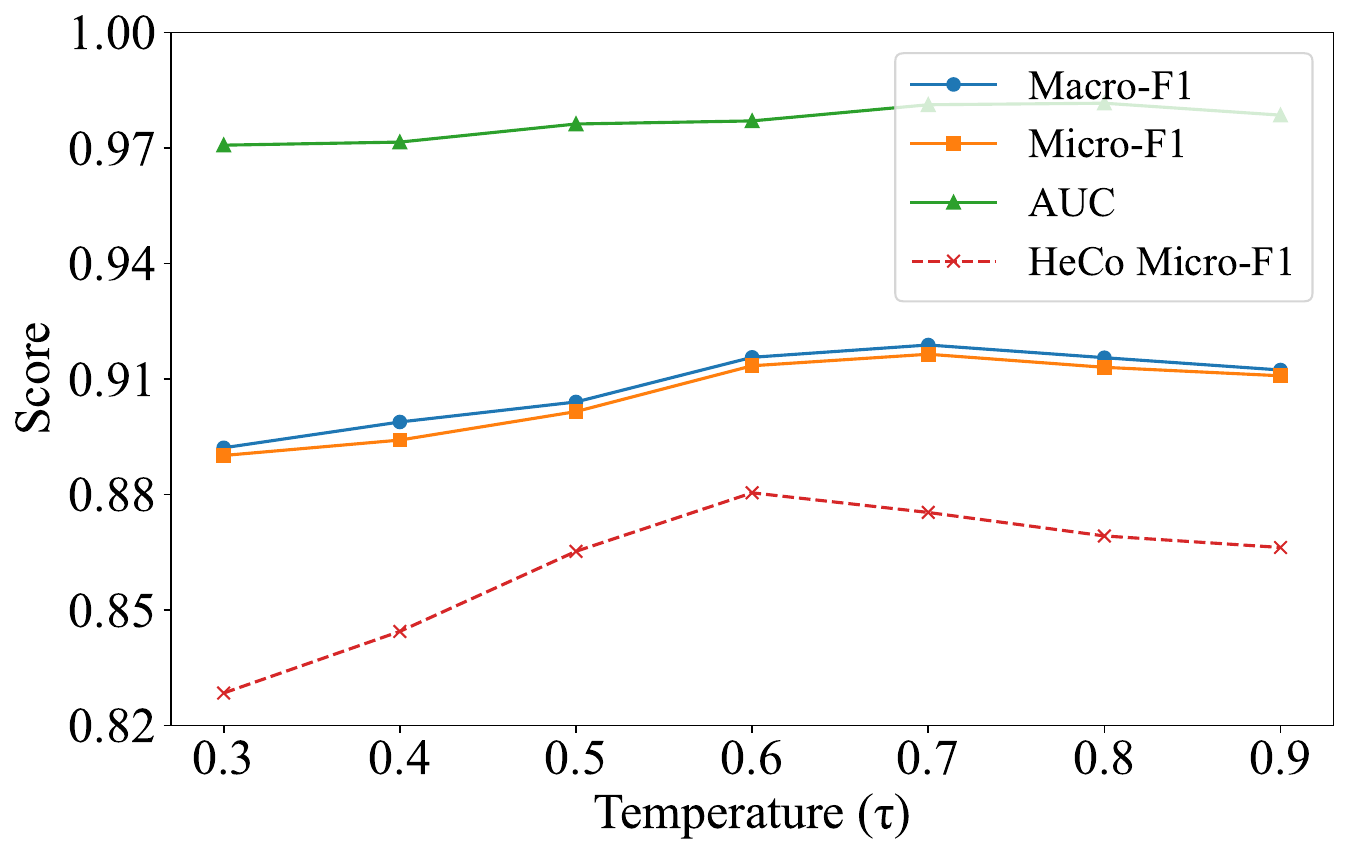}
        \caption{ACM}
        \label{tauparam_acm}
    \end{subfigure}
    \begin{subfigure}[b]{0.43\textwidth}
        \centering
        \includegraphics[width=\textwidth]{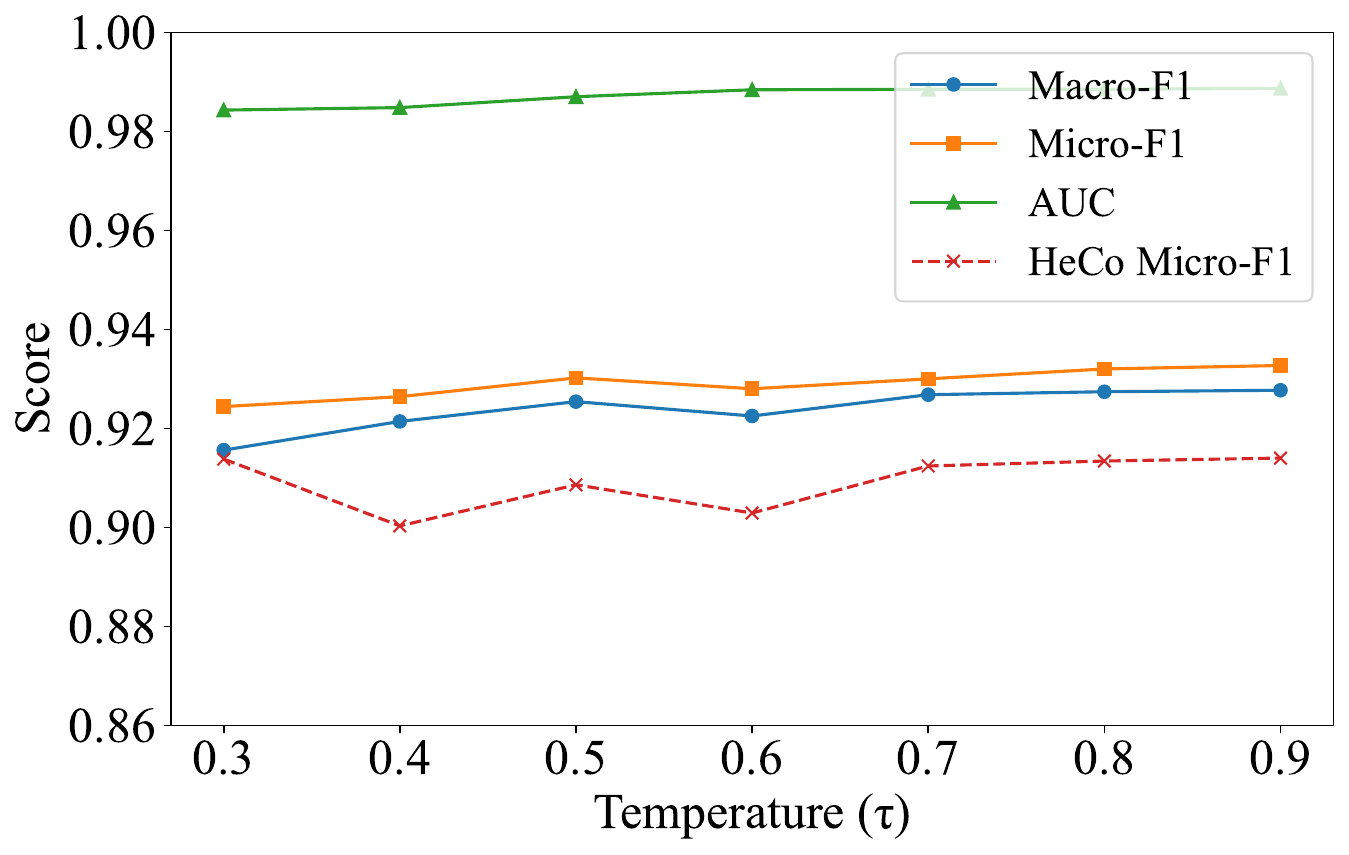}
        \caption{DBLP}
        \label{tauparam_dblp}
    \end{subfigure}
    \caption{Analysis of the temperature hyperparameter $\tau$ impact on model performance.}
    \label{tauparam}
\end{figure}

\textbf{Analysis of $\tau$.} The temperature hyperparameter $\tau$ controls the sharpness of the distribution in the contrastive learning objective. As illustrated in Figure~\ref{tauparam}, we systematically vary the temperature hyperparameter $\tau$ across a range of values to observe its effect on the model's performance. Our method demonstrates significantly less sensitivity to the temperature hyperparameter $\tau$ compared to HeCo. This robustness is particularly evident in the DBLP dataset, where our model maintains stable scores across various values of $\tau$. Furthermore, we observe that as the value of $\tau$ increases, the model's performance tends to stabilize, with scores exhibiting minimal variation for $\tau$ values between 0.6 and 0.9, indicating a high degree of consistency in this range.

\begin{figure}[!tb]
    \centering
    \begin{subfigure}[b]{0.43\textwidth}
        \centering
        \includegraphics[width=\textwidth]{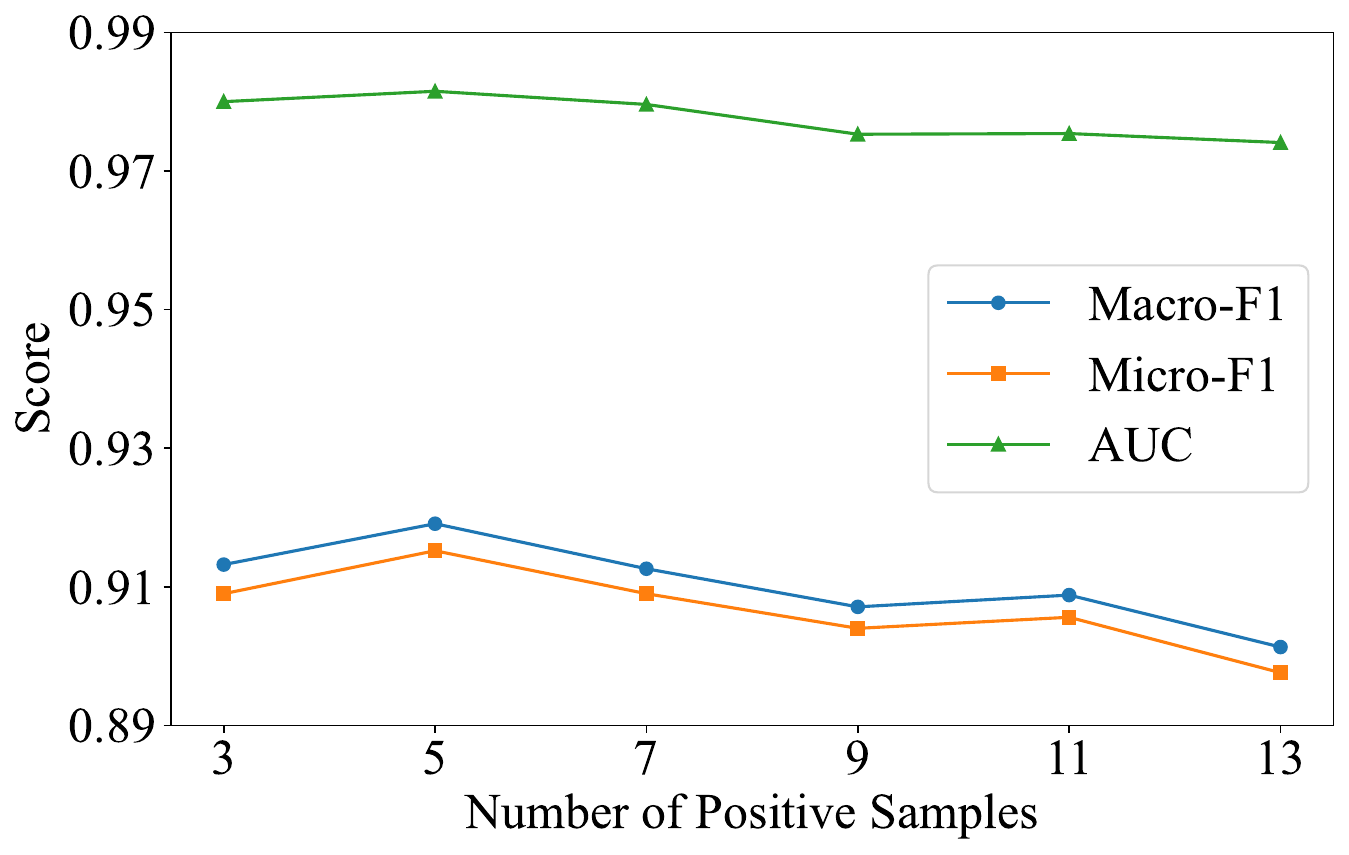}
        \caption{ACM}
        \label{parampos_acm}
    \end{subfigure}
    \begin{subfigure}[b]{0.43\textwidth}
        \centering
        \includegraphics[width=\textwidth]{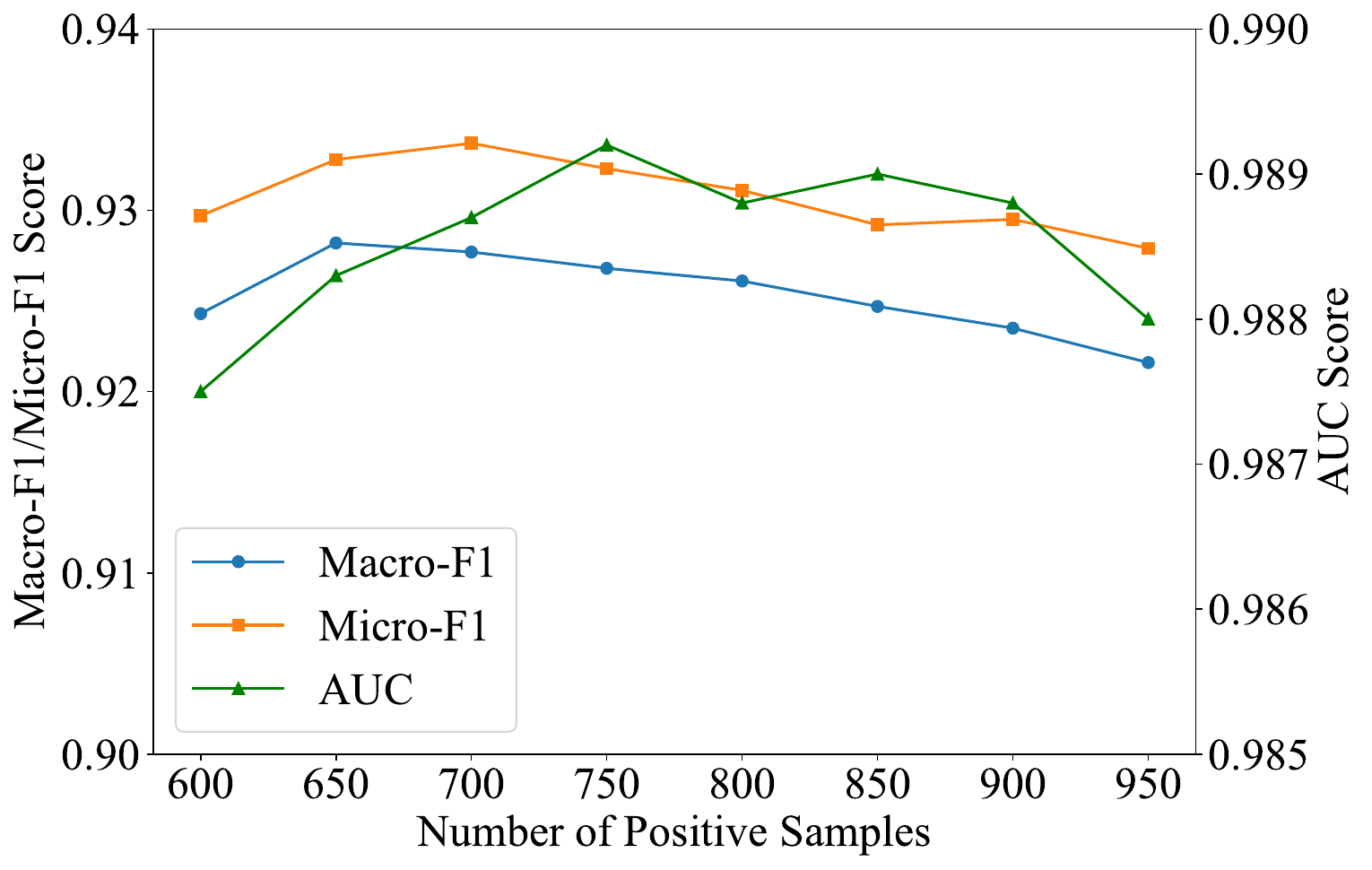}
        \caption{DBLP}
        \label{parampos_dblp}
    \end{subfigure}
    \caption{Analysis of the positive sample count $k$ impact on model performance.}
    \label{parampos}
\end{figure}

\textbf{Analysis of \(k\).}
The hyperparameter $k$ represents the number of sampled positive samples. As illustrated in Figure~\ref{parampos}, we observe the following trend in the ACM dataset. The model's performance initially increases as $k$ increases, reaching its peak when sampling five positive samples, then declines with further increases in $k$. The DBLP dataset demonstrates a similar trend, with optimal performance achieved at $k=700$. This higher optimal $k$ value for DBLP can be attributed to the dataset's characteristics, where nodes typically have a larger number of meta-path based neighbors, averaging around 1000. Sampling more positive samples helps capture a more comprehensive neighborhood structure, enhancing the richness and robustness of node representations.

\section{Conclusion}\label{co}

In this paper, we propose Incorporating Attributes and Multi-Scale Structures for Heterogeneous Graph Contrastive Learning (ASHGCL), a novel approach that constructs three distinct views using nodes' first-order neighbors, meta-path based neighbors, and node attributes. Through contrastive learning among these three views, our method simultaneously captures low-order structural information, high-order structural information, and node feature information, resulting in more comprehensive node representations. Furthermore, we introduce an innovative positive sampling strategy that selects similar positive samples for nodes based on both topological structure and node attributes, effectively mitigating sampling bias and improving the model's ability to distinguish semantically similar nodes.   Extensive experiments on various real-world datasets demonstrate the superiority of ASHGCL over state-of-the-art methods in node classification and clustering tasks.

\section*{Acknowledgements}
We would like to thank the anonymous reviewers for their time and effort in reviewing this manuscript. We will carefully consider their suggestions to improve the quality of the paper.





\bibliography{HG}

\end{document}